%% file: main_icml.tex
\documentclass{article}

\usepackage{hyperref}

\usepackage[accepted]{icml2023}

\input{setup/math_commands.tex}

\input{setup/packages}
\input{setup/acronyms}

\usepackage[textsize=tiny]{todonotes}

\icmltitlerunning{Unscented Autoencoder}
\begin{document}

\twocolumn[
\icmltitle{Unscented Autoencoder}



\icmlsetsymbol{equal}{*}

\begin{icmlauthorlist}
\icmlauthor{Faris Janjo\v{s}}{aaa}
\icmlauthor{Lars Rosenbaum}{aaa}
\icmlauthor{Maxim Dolgov}{aaa}
\icmlauthor{J. Marius Z\"ollner}{bbb}
\end{icmlauthorlist}

\icmlaffiliation{aaa}{Robert Bosch GmbH, Corporate Research, 71272 Renningen, Germany}
\icmlaffiliation{bbb}{Research Center for Information Technology (FZI), 76131 Karlsruhe, Germany}

\icmlcorrespondingauthor{}{first-name.last-name@de.bosch.com}

\icmlkeywords{Machine Learning, ICML}

\vskip 0.3in
]



\printAffiliationsAndNotice{}  

\input{chapters/abstract}


\input{chapters/introduction}
\input{chapters/related_work}
\input{chapters/problem_description}
\input{chapters/method}
\input{chapters/results}
\input{chapters/conclusion}

\newpage
\bibliography{bibliography/bibliography}
\bibliographystyle{icml2023}

\newpage
\appendix
\onecolumn
\section*{Appendix}
\input{chapters/appendix}

\end{document}

%% file: setup/math_commands.tex
\usepackage{amsmath,amsfonts,bm}

\def\eqref#1{equation~\ref{#1}}

\def\1{\bm{1}}

\def\rvx{{\mathbf{x}}}

\def\rvz{{\mathbf{z}}}

\def\vzero{{\bm{0}}}

\def\vmu{{\bm{\mu}}}

\def\vchi{{\bm{\chi}}}

\def\vf{{\bm{f}}}

\def\mL{{\bm{L}}}

\def\mSigma{{\bm{\Sigma}}}

\DeclareMathAlphabet{\mathsfit}{\encodingdefault}{\sfdefault}{m}{sl}
\SetMathAlphabet{\mathsfit}{bold}{\encodingdefault}{\sfdefault}{bx}{n}

\newcommand{\E}{\mathbb{E}}

\newcommand{\KL}{D_{\mathrm{KL}}}

\DeclareMathOperator*{\argmax}{arg\,max}
\DeclareMathOperator*{\argmin}{arg\,min}

%% file: setup/packages.tex
\usepackage{amsbsy}
\usepackage{mathtools}
\usepackage{amsmath}
\usepackage{amssymb}
\usepackage{amsthm}

\usepackage{hyperref}
\usepackage{url}
\usepackage[capitalize,noabbrev]{cleveref}

\usepackage{graphicx}
\usepackage{tabularx,booktabs}
\usepackage{colortbl}

\usepackage{subcaption}
\newcolumntype{Y}{>{\centering\arraybackslash}X}
\newcolumntype{b}{>{\hsize=.7\hsize}X}
\newcolumntype{s}{>{\hsize=.23\hsize}X}
\newcolumntype{a}{>{\hsize=.19\hsize}X}
\newcolumntype{t}{>{\hsize=.05\hsize}X}
\newcolumntype{m}{>{\hsize=.34\hsize}X}

\usepackage[utf8]{inputenc}
\usepackage{pgfplots}
\DeclareUnicodeCharacter{2212}{−}
\usepgfplotslibrary{groupplots,dateplot}
\usetikzlibrary{patterns,shapes.arrows}
\pgfplotsset{compat=newest}

\usepackage[nolist]{acronym}
\usepackage{mfirstuc} 
\MFUnocap{the}\MFUnocap{and}\MFUnocap{of}\MFUnocap{with}\MFUnocap{for}\MFUnocap{a}\MFUnocap{by}\MFUnocap{to}

\theoremstyle{plain}

\theoremstyle{definition}

\theoremstyle{remark}

%% file: setup/acronyms.tex
\begin{acronym}
	\acro{VAE}{Variational Autoencoder}
	\acro{FID}{Fréchet Inception Distance}
	\acro{UAE}{Unscented Autoencoder}
	\acro{RAE}{Regularized Autoencoder}
	\acro{WAE}{Wasserstein Autoencoder}
	\acro{IWAE}{Importance-Weighted Autoencoder}
	\acro{ELBO}{Evidence Lower Bound}
	\acro{UKF}{Unscented Kalman Filter}
	\acro{EKF}{Extended Kalman Filter}
	\acro{GMM}{Gaussian Mixture Model}
	\acro{UT}{Unscented Transform}
	\acro{KL}{Kullback-Leibler}
	\acro{PCA}{Principal Component Analysis}
	\acro{SVD}{Singular Value Decomposition}
	\acro{CV}{Coefficient of Variation}
\end{acronym}

%% file: chapters/abstract.tex
\begin{abstract}
The \ac{VAE} is a seminal approach in deep generative modeling with latent variables. Interpreting its reconstruction process as a nonlinear transformation of samples from the latent posterior distribution, we apply the \ac{UT} -- a well-known distribution approximation used in the \ac{UKF} from the field of filtering. A finite set of statistics called sigma points, sampled deterministically, provides a more informative and lower-variance posterior representation than the ubiquitous noise-scaling of the reparameterization trick, while ensuring higher-quality reconstruction. We further boost the performance by replacing the \ac{KL} divergence with the Wasserstein distribution metric that allows for a sharper posterior. Inspired by the two components, we derive a novel, deterministic-sampling flavor of the \ac{VAE}, the \ac{UAE}, trained purely with regularization-like terms on the per-sample posterior. We empirically show competitive performance in \ac{FID} scores over closely-related models, in addition to a lower training variance than the \ac{VAE}\footnote{Code available at: \url{https://github.com/boschresearch/unscented-autoencoder}}.
\end{abstract}

%% file: chapters/introduction.tex
\section{Introduction}
\begin{figure}[h!]
	\centering
	\includegraphics[width=0.85\linewidth]{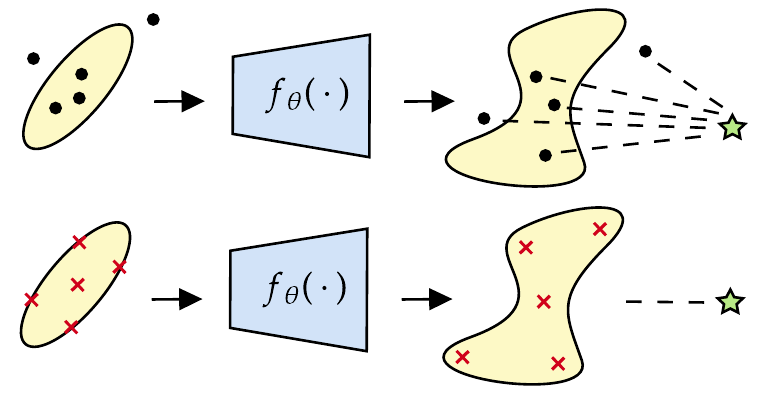}
	\small\caption{The VAE decoder $f_\theta(\cdot)$ can be interpreted as a nonlinear mapping of the Gaussian posterior distribution generated by the encoder, resulting in a non-Gaussian output distribution. The standard \ac{VAE} (top) samples randomly from the posterior (black points) and matches each decoded sample to the ground truth (green star). Our model (bottom) samples and transforms fixed posterior sigma points (red) instead. By matching the mean of the transformed points, we push the entire output distribution to resemble the ground truth.}
	\label{fig:illustration}
\end{figure}

The Variational Autoencoder (VAE)~\citep{rezende2014variational,kingma2015variational} is a widely used method for learning deep latent variable models via maximization of the data likelihood using a reparametrized version of the \ac{ELBO}. Deep latent variable models are used as generative models in a variety of application domains such as image~\citep{vahdat2020NVAE}, language~\citep{bowman2015generating,kusner2017grammar}, and dynamics modeling~\citep{karl2016deep}. A good generative model requires the \ac{VAE} to produce high-quality samples from the prior latent variable distribution and a disentangled latent representation is desired to control the generation process~\citep{higgins2017betavae}. Another important application of deep latent variable models is representation learning, where the goal is to induce a latent representation facilitating downstream tasks~\citep{bengio2013replearning,townsend2019compression,tripp2020lso,rombach2022high}. In many of these tasks a good sample quality, as well as a 'well-behaved' latent representation with a high reconstruction accuracy is desired.

Since their introduction, \ac{VAE}s have been one of the methods of choice in generative modeling due to their comparatively easy training and the ability to map data to a lower dimensional representation as opposed to generative adversarial networks~\citep{goodfellow2014gan}. However, despite their popularity there are still open challenges in \ac{VAE} training addressed by recent works. A major problem of \ac{VAE}s is their tendency to have a trade-off between the quality of samples from the prior and the reconstruction quality. This trade-off can be attributed to overly simplistic priors~\citep{bauer2019resampled}, encoder/decoder variance~\citep{dai2018diagnosing}, weighting of the \ac{KL} divergence regularization ~\citep{higgins2017betavae, tolstikhin2018wae}, or the aggregated posterior not matching the prior~\citep{tolstikhin2018wae,ghosh2019variational}. Furthermore, the \ac{VAE} objective can be prone to spurious local maxima leading to posterior collapse~\citep{chen2017postcollapse,lucas2019postcollapse,dai2020usual}, which is characterized by the latent posterior (partially) reducing to an uninformative prior. Finally, the variational objective requires approximations of expectations by sampling, which causes increased gradient variance~\citep{burda2016iwae} and makes the training sensitive to several hyperparameters~\citep{bowman2015generating,higgins2017betavae}.

Our main technical contributions are two modifications to the original \ac{VAE} objective resulting in an improved sample and reconstruction quality. We propose to use a well-known algorithm from the filtering and control literature, the Unscented Transform (UT)~\citep{uhlmann1995dynamic}, to obtain lower-variance, albeit potentially biased gradient estimates for the optimization of the variational objective. A lower variance is achieved by only sampling at the sigma points of the variational posterior and transforming these points with a deterministic decoder. In this context, we show that reconstructing the entire posterior distribution via its sigma points (visualized in Fig.~\ref{fig:illustration}) is superior in resulting image quality to reconstructing individual random samples. Furthermore, we observe that the regularization toward a standard normal prior using a \ac{KL} divergence often harshly penalizes low variance along some components even though the low variance is usually beneficial for reconstruction. Thus, we use a different regularization based on the Wasserstein metric~\citep{patrini2020sinkhorn}. To account for resulting sharper posteriors, we add a regularizer for decoder smoothness around the mean encoded value, similar to~\cite{ghosh2019variational}. We conduct rigorous experiments on several standard image datasets to compare our modifications against the \ac{VAE} baseline, the closely-related \ac{RAE}~\citep{ghosh2019variational}, the \ac{IWAE}~\citep{burda2016iwae}, as well as the \ac{WAE}~\citep{tolstikhin2018wae}. 

%% file: chapters/related_work.tex
\section{Related Work}
Many recent works on \ac{VAE}s focus on understanding and addressing still existing problems like undesired posterior collapse~\citep{dai2020usual}, trade-off between sample and reconstruction quality~\citep{tolstikhin2018wae, bauer2019resampled}, or non-interpretable latent representations~\citep{rolinek2019variational, higgins2017betavae}. Other recent works suggest to move from the probabilistic \ac{VAE} models to deterministic models, such as the \ac{RAE} in~\cite{ghosh2019variational}; our model can be considered as part of this class. As previously mentioned, we employ two major modifications to the \ac{VAE}, namely the \textbf{Unscented Transform} and the \textbf{Wasserstein metric}, as well as \textbf{decoder regularization}; we outline the section accordingly.

We use the \textbf{Unscented Transform}~\citep{uhlmann1995dynamic} from the field of nonlinear filtering within signal processing. In this context, the signal state estimate is often assumed to be Gaussian in order to maintain tractability. However, nonlinear prediction and measurement models always invalidate this assumption at each time step so that a re-approximation becomes necessary. A commonly used approach is the \ac{EKF}, where a linearization of the models is employed so that the Gaussian state remains Gaussian during filtering. In contrast, alternative approaches that represent the Gaussian state (assuming application in the context of the \ac{VAE} posterior) with samples for propagation and update have emerged. These approaches can be clustered according to the employed sampling method -- random as in (Gaussian) particle filters~\citep{Doucet2008ATO} or deterministic, e.g. in the \ac{UKF}~\citep{julier2000new}. In the \ac{UKF}, the $n$-dimensional Gaussian is approximated with $2n+1$ deterministic samples, which can be propagated through the nonlinearities and are sufficient for computing the statistics of a Gaussian distribution, i.e. its mean and covariance. This procedure is referred to as the Unscented Transform (UT). 

The use of deterministic sampling\footnote{Sampling from a set of points at fixed locations in the domain.} aims to achieve a good coverage of the distribution represented with the mean and covariance. Although this approach produces biased estimates of the involved expectations compared to random sampling due to non-i.i.d. samples, it often captures well the nonlinearities applied to the distribution, for a finite, small set of samples in the filtering context. This observation can transfer to neural networks due to their Lipschitz continuity~\cite{khromov2023fundamental}. Our \ac{UT} experiments empirically underline this expectation. For a more comprehensive overview of the \ac{UT} and the \ac{UKF}, we refer the reader to~\cite{menegaz2015systematization}.

The \ac{UT} uses several samples to get an estimate of the moments of a nonlinearly transformed probability distribution. Along those lines, our method also relates to the \ac{IWAE}~\citep{burda2016iwae} and some of its extensions~\citep{tucker2018doubly}. \ac{IWAE} uses importance weighting of $K$ posterior samples to obtain a variational distribution closer to the true posterior~\citep{cremer2017reinterpretingiwae}. The method is known to have a diminishing gradient signal for the inference network~\citep{rainforth2018tighter} if no additional improvements are used~\citep{tucker2018doubly}. Using the Wasserstein metric, the inference distribution is sharp, so practically there is not much gain in a more complex distribution. However, multiple samples can help to obtain lower variance gradient estimates, which also applies to the \ac{IWAE} by taking a multiple of $K$ samples. Sampling only at the sigma points reduces this variance even more and is known to empirically work well in filtering and control.

The \textbf{Wasserstein metric} is used in~\cite{tolstikhin2018wae,patrini2020sinkhorn} to regularize the aggregated posterior $q_{\text{agg}}(\rvz)=\E_{p(\rvx)}\left[ q(\rvz|\rvx)\right]$ toward the standard normal prior. The authors also show that such an objective is an upper bound to the Wasserstein distance between the sampling distribution of the generative model and the data distribution if the regularization is scaled by the Lipschitz constant of the generator. In contrast, we do not regularize the aggregated posterior, but use the Wasserstein distance to weakly regularize the mean and variance of the encoder, such that neither explodes and we can do ex-post density estimation. From a theoretical point of view, we do not fix the prior but learn the manifold; the aggregated posterior is learned by fitting a mixture to the encoded data points.


Finally, our work incorporates several ideas from the recently published \ac{RAE}~\citep{ghosh2019variational}. We also use a \textbf{decoder regularization} term based on the decoder Jacobian in our loss, which promotes smoothness of the latent space. In contrast to the \ac{RAE} however, we generalize the term from a deterministic to a stochastic encoder as not every data point might be encoded with the same fidelity. Furthermore, we employ ex-post density estimation as we do not explicitly regularize the aggregated posterior toward a prior. Conceptually, the \ac{UAE} can be placed between the \ac{VAE}, characterized by significant sampling variance, and the purely deterministic \ac{RAE}.


%% file: chapters/problem_description.tex
\section{Problem Description}\label{sec:problem_desc}
Most generative models take a max-likelihood approach to model a real-world distribution~$p(\rvx)$ via the $\theta$-parameterized probabilistic generator model $p_\theta(\rvx)$
\begin{equation}\label{eq:max_likelihood}
\theta \leftarrow {\argmax}_\theta \quad \E_{\rvx\sim p(\rvx)} [\log p_\theta(\rvx)]\ .
\end{equation}
In this setting, latent variable generative approaches assume an underlying structure in $p(\rvx)$ not directly observable from the data and model this structure with a latent variable $\rvz$, which is well-motivated by de Finetti's theorem~\citep{accardi2001finetti}. As a result, the distribution $p(\rvx)$ can be represented as a product of tractable distributions. However, directly incorporating $\rvz$ via an integral $\int p_\theta(\rvx|\rvz)p(\rvz)d\rvz$ is intractable; thus, one introduces an amortized variational distribution $q_\phi(\rvz|\rvx)$~\citep{zhang2018advances} and obtains
\begin{equation}\label{eq:latent_objective}
\log p_\theta(\rvx) = \log \E_{\rvz\sim q_\phi(\rvz|\rvx)} \left[ \frac{p_\theta(\rvx,\rvz)}{q_\phi(\rvz|\rvx)} \right] \ .
\end{equation}
This model assumption is the basis of variational inference. Applying Jensen's inequality yields the well-known \ac{ELBO}, denoted by $\mathcal{L}$ 
\begin{align}\label{eq:elbo}
\begin{split}
\log p_\theta(\rvx) \geq \mathcal{L} &= \E_{\rvz\sim q_\phi(\rvz|\rvx)}[\log p_\theta(\rvx|\rvz)] - \\
&- \KL(q_\phi (\rvz|\rvx) \Vert p(\rvz))\ ,
\end{split}
\end{align}
which is maximized w.r.t. $\theta$ and $\phi$. The first term accounts for the quality of reconstructed samples and the $\KL(...)$ term pushes the approximate posterior to mimic the prior, i.e. it enforces a $p(\rvz)$-like structure to the latent space.

Training on $\mathcal{L}$ in Eq.~(\ref{eq:elbo}) requires computing gradients w.r.t. $\theta$ and $\phi$. This is relatively straightforward for the generator parameters, however, requiring a high-variance policy gradient for the posterior parameters. To avoid this issue in practice, the reparameterization trick~\citep{kingma2015variational} is used to simplify the sampling of the approximate posterior by means of an easy-to-sample distribution. Assuming a Gaussian posterior  $\mathcal{N}(\vmu, \mSigma)$, we can sample a multivariate normal and obtain the latent feature vector via the deterministic transformation
\begin{equation}\label{eq:reparam}
\rvz = \vmu + \mL\bm{\epsilon}, \quad \bm{\epsilon}\sim\mathcal{N}(\mathbf{0}, \mathbf{I}),\quad\mSigma=\mL\mL^T\ .
\end{equation}

With the help of the reparameterization trick, the \ac{VAE}~\citep{kingma2013auto} provides a framework for optimizing the loss function from the condition in Eq.~(\ref{eq:elbo}) via an encoder--decoder generative latent variable model. The encoder $E_\phi(\rvx) = \{\bm{\mu}_\phi(\rvx), \bm{\Sigma}_\phi(\rvx)\}$ parameterizes a multivariate Gaussian $q_\phi(\rvz|\rvx)=\mathcal{N}(\rvz|\bm{\mu}_\phi(\rvx), \bm{\Sigma}_\phi(\rvx))$, where $\mSigma_\phi$ is usually a diagonal matrix, $\bm{\Sigma}_\phi=\text{diag}(\bm{\sigma}_\phi)$. The decoder $D_\theta(\rvz) = \bm{\mu}_\theta(\rvz)$ is in practice rendered deterministic: $p_\theta(\rvx|\rvz) = \mathcal{N}(\rvx|\bm{\mu}_\theta(\rvz), \mathbf{0})$, reducing the reconstruction term in Eq.~(\ref{eq:elbo}) to a simple mean-squared error under the expectation of the posterior~${\E_{\rvz\sim q_\phi(\rvz|\rvx)}\Vert\rvx-\bm{\mu}_\theta(\rvz)\Vert^2_2}$. The VAE uses the reparameterization trick for efficient sampling from the posterior $q_\phi$ (in practice providing only a single sample to the decoder), which enables a lower-variance gradient backpropagation through the encoder.

The deterministic decoder and the reparameterization trick allow for a slightly different interpretation of the reconstruction/generation process: a (highly) nonlinear transformation of an input distribution, represented (usually) only by a single stochastic sample. The sample is white noise\footnote{The white-noise interpretation is used in~\cite{ghosh2019variational} to justify regularization as an alternative to the noise sampling.}, scaled and shifted by the posterior moments. This interpretation serves as the basis for our work, where the unscented transform of the input distribution serves as an alternative to the single-stochastic-sample representation. In the next section, we outline the unscented transform representation of the input to the decoder via a set of deterministically computed and sampled sigma points.

%% file: chapters/method.tex
\section{Unscented Transform of the Posterior}
\subsection{Background}

\begin{figure*}[h!]
	\centering
	\begin{subfigure}[b]{0.55\columnwidth}
		\centering
		\includegraphics[width=1\linewidth,trim={1cm 1cm 1cm 1cm},clip]{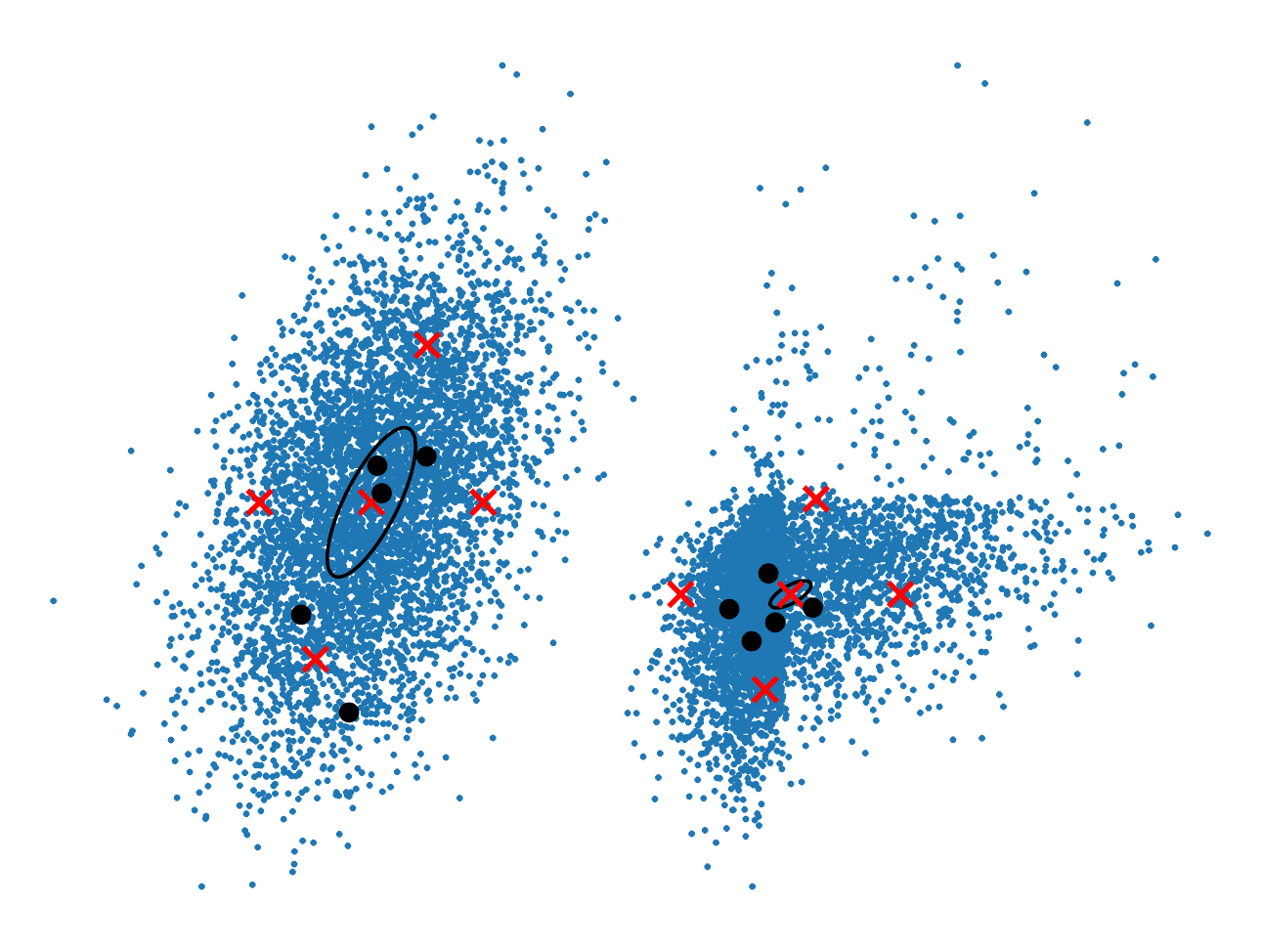}
		\caption{}
		\label{fig:sigma}
	\end{subfigure}
	\begin{subfigure}[b]{0.45\columnwidth}
		\centering
		\includegraphics[width=0.9\linewidth,trim={5cm 3cm 5cm 3cm},clip]{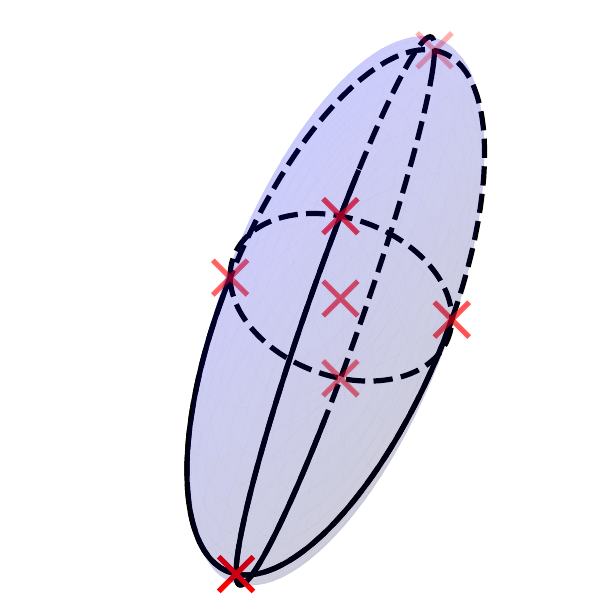}
		\caption{}
		\label{fig:sigma_ellipse}
	\end{subfigure}
	\begin{subfigure}[b]{0.55\columnwidth}
		\centering
		\includegraphics[width=1.2\linewidth,trim={0cm 0.5cm 0cm 1cm},clip]{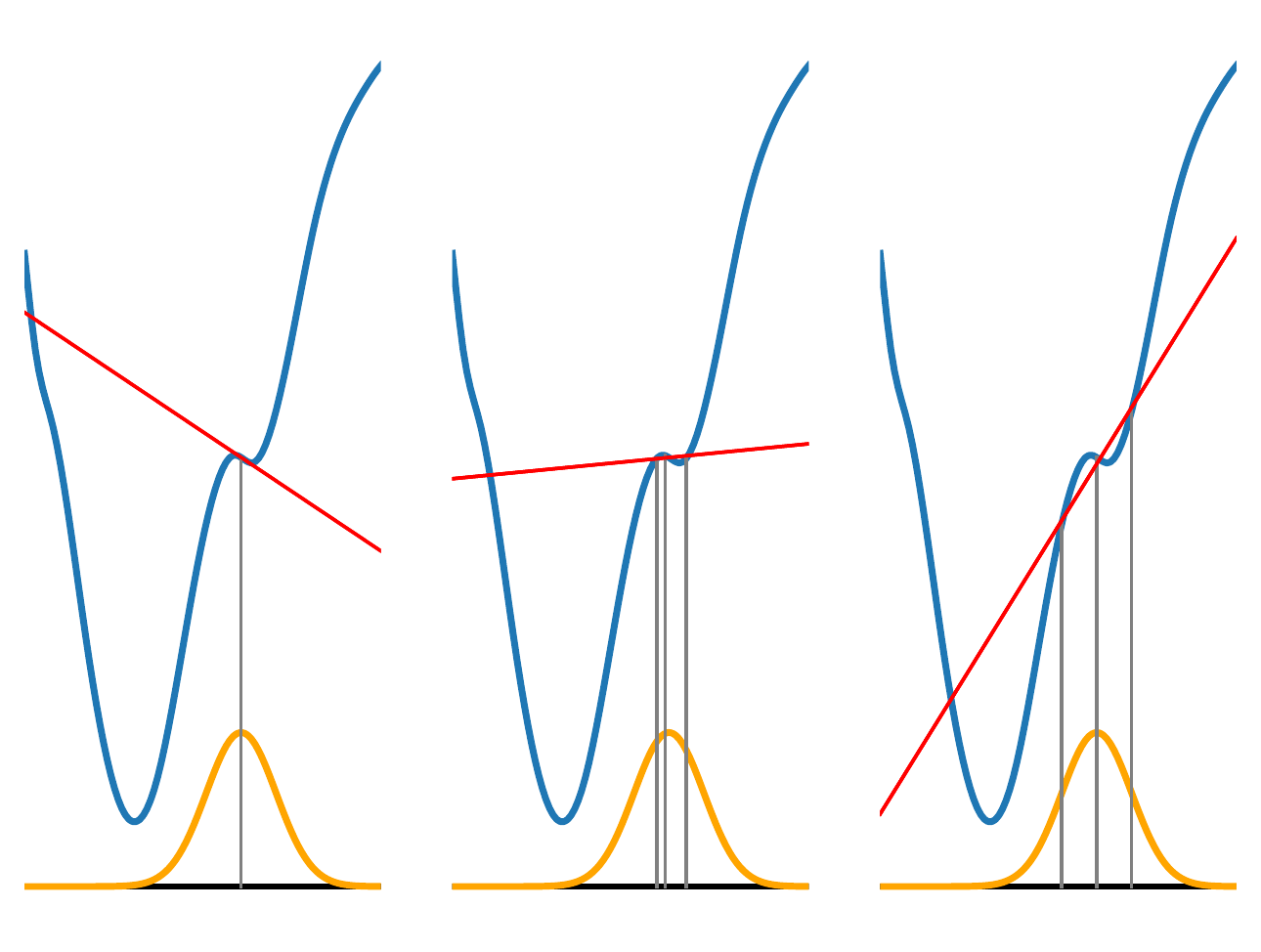}
		\caption{}
		\label{fig:sigma_gradient}
	\end{subfigure}\hfill
	\small\caption{(best viewed in color) (a) \textbf{(transforming 2D sigma points)} Left: a Gaussian with its Monte Carlo approximation (blue), sigma points computed according to Eq.~(\ref{eq:sigma}) (red), and five random samples (black points). Right: nonlinear RReLU activation \cite{xu2015empirical} applied to the distribution, sigma points, and the random samples. In this example, the five sigma points provide a better approximation of the transformed distribution than the five random samples. \\
		(b) \textbf{(3D sigma points)} Sigma points (red) on an ellipsoid spanned by a $3\times3$ covariance matrix, consisting of a central sigma point and a pair of sigma points on each axis.\\
		(c) \textbf{(gradient variance)} Left: loss function (blue) at a sample (gray) corresponding to the standard normal (yellow) mean. The gradient of the loss function (red) at the mean is not representative of the true gradient. Middle: a high-variance gradient computed from the gradients at the three random samples drawn from the standard normal, potentially far away from the true gradient. Right: gradient of the loss function computed from the gradients at the three sigma points; although the estimate is potentially biased due to the applied nonlinear transformation, it has lower variance than if computed from the random points. The three provided examples can be interpreted as the \ac{RAE}-\citep{ghosh2019variational}, \ac{VAE}-, and \ac{UAE}-like sampling procedures. 
	}
\end{figure*}

The unscented transform \cite{uhlmann1995dynamic} is a method to evaluate a nonlinear transformation of a distribution characterized by its first two moments. Assume a known deterministic function $\vf$ applied to a distribution $P(\vmu, \mSigma)$ with mean and covariance $\vmu\in\mathbb{R}^n$ and $\mSigma\in\mathbb{R}^{n\times n}$. If $\vf$ is a linear transformation, one can describe the distribution $Q(\hat{\vmu}, \hat{\mSigma})$ at the output via $\hat{\vmu}=\vf\vmu$ and $\hat{\mSigma}=\vf\mSigma\vf^T$. Similarly, for a nonlinear transformation $\vf$ but a zero covariance matrix $\mSigma=\mathbf{0}$, the mean of the transformed distribution is $\hat{\vmu}=\vf(\vmu)$. However, in the general case it is not possible to determine $\hat{\vmu}$ and $\hat{\mSigma}$ of the $\vf$-transformed distribution given $\vmu$ and $\mSigma$ since the result depends on higher-order moments. Thus, the unscented transform is useful; it provides a mechanism to obtain this result via an approximation of the input distribution while assuming full knowledge of $\vf$.  

In computing the unscented transform, first a set of sigma points characterizing the input~$P(\vmu, \mSigma)$ is chosen. The most common approach \cite{menegaz2015systematization} is to take a set $\{\vchi_i\}^{2n}_{i=0}$, $\vchi_i\in\mathbb{R}^n$ of $2n+1$ symmetric points centered around the mean (incl. the mean), e.g. for $1\leq i \leq n$,
\begin{align}\label{eq:sigma}
\begin{split}
\vchi_0 &= \vmu\ , \\
\vchi_{i} &= \vmu + \sqrt{(\kappa + n)\mSigma} \big{|}_i\ , \\
\vchi_{i+n} &=  \vmu - \sqrt{(\kappa + n)\mSigma} \big{|}_i \ ,
\end{split}
\end{align}
where $\kappa>-n$ is a real constant and $\big{|}_i$ denotes the $i$-th column. The approximation in Eq.~(\ref{eq:sigma}) is unbiased; the mean and covariance of the sigma points are $\vmu$ and $\mSigma$. Thus, one can compute the transformation $\hat{\vchi}_i=\vf(\vchi_i)$ and estimate the mean and covariance of the $\vf$-transformed distribution
\begin{align}
\hat{\vmu} &= \textstyle \frac{1}{2n+1} \sum^{2n}_{i=0} \hat{\vchi}_i\ , \label{eq:sigma_mu_transformation} \\
\hat{\mSigma} &= \textstyle \frac{1}{2n+1} \sum^{2n}_{i=0} (\hat{\vchi}_i - \hat{\vmu})(\hat{\vchi}_i - \hat{\vmu})^T \ . \label{eq:sigma_cov_transformation}
\end{align}
A visualization of the sigma points and their transformation is depicted in Fig. \ref{fig:sigma}. The procedure in Eq.~(\ref{eq:sigma}-\ref{eq:sigma_cov_transformation}) effectively applies the fully-known function $\vf$ to an approximating set of points whose mean and covariance equal the original distribution's. Therefore, in the context of the commonly used VAE decoder nonlinearities, the mean and covariance of the transformed sigma points can be closer to the true transformed mean and covariance compared to the ones computed by propagating the same number of random samples from the original distribution.

\subsection{Unscented Transform in the VAE}
In an \ac{ELBO} maximization setting from Eq.~(\ref{eq:elbo}), the nonlinear transformation of the posterior in the decoder lends itself straightforwardly to the unscented transform approximation. Given any posterior defined by $\vmu$ and $\mSigma$, we can compute the sigma points (for example according to Eq.~(\ref{eq:sigma})) and provide them to the decoder. In a \ac{VAE}, the sigma points provide a deterministic-sampling alternative to the reparameterization-trick-computed random samples of the latent space. Furthermore, computing the average reconstruction of the sigma points at the output of the decoder provides an approximation of the mean of the entire transformed posterior distribution in Eq.~(\ref{eq:sigma_mu_transformation}), while implicitly taking into account the variance in Eq.~(\ref{eq:sigma_cov_transformation}), as opposed to the per-sample reconstructions.

The choice of the number of sigma points provided to the decoder is similar to the sampling in Eq.~(\ref{eq:reparam}), where one can realize a single latent vector with a single sample from $\mathcal{N}(\vzero, \mathbf{I})$ or multiple latents, resulting in a trade-off between reconstruction quality and computation demands \cite{ghosh2019variational}. However, taking a single or few random samples in the \ac{VAE} setting can produce instances very far from the mean, especially in high dimensional spaces. In contrast, sampling sigma points produces a more controlled overall estimate of the posterior (as well as producing a more accurate transformed posterior, see Eq. (\ref{eq:sigma_mu_transformation}-\ref{eq:sigma_cov_transformation})) since the samples lie on the border of a hyperellipsoid induced by the covariance matrix $\mSigma$ (example in Fig. \ref{fig:sigma_ellipse}). Thus, while computing the loss function gradients (which are a function of the samples), the sigma-sampling has the potential to bring a more accurate and lower-variance estimate when all the sigma points are considered. This is illustrated in Fig \ref{fig:sigma_gradient}. Further empirical arguments validating the lower gradient variance claim are provided in Appendix \ref{sec:grad_variance}.

The sigma-sampling of the \ac{UT} can be applied to any learned posterior described by its first two moments (as common in generative models), not only the \ac{VAE} standard normal. With this description, the sigma points cannot be the uniquely optimal representation of the distribution since there is an infinite number of distributions that share the first two moments. However, the \ac{UT} has shown superior empirical performance over other representations in extensive experiments in~\cite{julier2000new} and~\cite{zhang2009accuracy}, under various distributions and nonlinear functions, and especially for the case of differentiable functions. This has led to the \ac{UKF}, built on this paradigm, being one of the major algorithms in filtering and control. Guided by the success of the method, we hypothesize that applying the \ac{UT} in the \ac{VAE} setting has the potential to, for a finite set of samples, provide a better approximation of the learned two-moment Gaussian posterior than the ubiquitous independent random sampling and reconstruction. With these insights, we develop the \ac{UAE} model presented in the next section.

\section{Unscented Autoencoder (UAE)}
The \ac{UAE} is a deterministic-sampling autoencoder model maximizing the \ac{ELBO}. It addresses the maximum likelihood optimization problem from Sec.~\ref{sec:problem_desc}, namely the $\mathcal{L}$ maximization from Eq.~(\ref{eq:elbo}), by computing the \ac{UT} of the posterior $q_\phi(\rvz|\rvx)$ parameterized by the encoder $E_\phi(\rvx) = \{\vmu_\phi(\rvx), \mSigma_\phi(\rvx)\}$ (see Eq.~(\ref{eq:sigma}-\ref{eq:sigma_cov_transformation})). The latent features $\rvz$ can be obtained by deterministically sampling multiple sigma points, resulting in a lower variance sampling than of the reparameterization trick in Eq.~(\ref{eq:reparam}). Good performance of the model is further boosted by replacing the vanilla KL divergence with the Wasserstein distribution metric, which effectively performs a regularization of the posterior moments. The decoder regularization applies an additional smoothing effect on the latent space -- it is formally derived in Sec.~\ref{subsec:elbo_derivation}. The full training objective consists of optimizing $\phi, \theta \leftarrow \argmin_{\phi, \theta} \mathcal{L}_{\text{UAE}}$,
\begin{equation}\label{eq:uae_obj}
\mathcal{L}_{\text{UAE}} = \E_{\rvx\sim p_\text{data}} \mathcal{L}_{\text{REC}} +  \beta\mathcal{L}_{W} + \gamma\mathcal{L}_{D_\theta\text{REG}}\ ,
\end{equation}
where $\beta$ (from the $\beta$-\ac{VAE}~\cite{higgins2017betavae}) and $\gamma$ are weights.

The \textbf{reconstruction term} $\mathcal{L}_{\text{REC}}$ is an $L_2$ loss function incorporating the average of decoded sigma points
\begin{align}\label{eq:rec_loss}
\begin{split}
\mathcal{L}_{\text{REC}} &= \Vert \rvx - \textstyle \frac{1}{K}\sum^K_{k=1} D_\theta(\rvz_k)\Vert^2_2\ , \\ \quad \rvz_k&\sim \{\vchi_i(\vmu_\phi, \mSigma_\phi)\}^{2n}_{i=0}\ ,
\end{split}
\end{align}
where $K$ $n$-dimensional vectors $\rvz_k$ are sampled from the set of sigma points, $K\leq 2n+1$. Various sampling heuristics are investigated in Appendix~\ref{sec:multi_sigma}. Note that this reconstruction loss function differs from the commonly used~$\textstyle\frac{1}{K}\sum^K_{k=1}  \Vert \rvx - D_\theta(\rvz_k)\Vert^2_2$, where each decoded sample is matched to the ground truth. This strategy, employed in the standard multi-sample VAE, aims at getting the same output image for different samples thus demanding a certain attenuation property from the deterministic decoder. In contrast, Eq.~(\ref{eq:rec_loss}) is motivated by the application of the \ac{UT} in filtering where after propagating the sigma points through a nonlinear function a Gaussian is fit to the posterior (see Eq.~(\ref{eq:sigma_mu_transformation}-\ref{eq:sigma_cov_transformation})). By applying the loss to the mean output image, we essentially maintain a probability distribution at the output.

We use the \textbf{Wasserstein metric term} $\mathcal{L}_{\text{W}}$ as an alternative to the KL divergence. For a multivariate posterior and a multivariate normal prior, the KL divergence is defined as
\begin{align}\label{eq:kl_div_log_l}
\mathcal{L}_{\text{KL}} = \Vert\vmu_\phi\Vert^2_2 + \text{tr}(\mSigma_\phi) - n - 2\text{tr}(\log\bm{L}_\phi)\ ,
\end{align}
in the general case\footnote{Derived from $\KL(\mathcal{N}_0\Vert \mathcal{N}_1) = \frac{1}{2}(\text{tr}(\mSigma^{-1}_1\mSigma_0) - n + (\vmu_1 - \vmu_0)^T\mSigma^{-1}_1(\vmu_1-\vmu_0) + \log(\frac{\det \mSigma_1}{\det \mSigma_0}))$ for $\mathcal{N}_1(\mathbf{0}, \mathbf{I})$ and $\mSigma=\bm{L}\bm{L}^T$.} of a full-covariance matrix $\mSigma_\phi=\bm{L}_\phi\bm{L}^T_\phi$. Instead, due to favorable optimization properties and higher-quality reconstruction, we use the Wasserstein metric between distributions. This metric effectively replaces the covariance part of the KL term, $\text{tr}(\mSigma_\phi) - 2\text{tr}(\log\bm{L}_\phi)$, with the squared Frobenius norm of the mismatch between the lower triangular matrix and the identity
\begin{equation}\label{eq:frob_approx}
	\mathcal{L}_W = \Vert \bm{L}_\phi - \mathbf{I} \Vert^2_F = \text{tr}(\mSigma_\phi) - 2\text{tr}(\bm{L}_\phi) \ .
\end{equation}
It differs from the original objective in Eq.~(\ref{eq:kl_div_log_l}) only in the lack of a logarithm while sharing the same global minimum. Further details are provided in Sec. \ref{subsec:kl_div_approx}. Such a loss function allows the variance to approach zero (which is instead strongly penalized by the logarithm in Eq.~(\ref{eq:kl_div_log_l})), yielding a sharper posterior.

The \textbf{decoder regularization term} $\mathcal{L}_{D_\theta\text{REG}}$ is a generalization of the gradient penalty term in~\cite{ghosh2019variational}, accounting for a fully probabilistic formulation. It can be realized as a penalty on the input--output gradient of the posterior mean, weighted by the largest eigenvalue of the covariance matrix
\begin{equation}\label{eq:uae_dec_reg}
\mathcal{L}_{D_\theta\text{REG}} = \lambda_{\max}(\mSigma_\phi) \Vert \nabla_{\vmu_\phi} D_\theta(\vmu_\phi)\Vert^2_2\ .
\end{equation}
We approximate $\lambda_{\max}(\mSigma_\phi)$ by the largest diagonal, which is correct for a diagonal $\mSigma_\phi$.

We provide an overview of the \ac{VAE}, \ac{RAE}, and \ac{UAE} loss functions in Tab. \ref{tab:loss_functions}, together with the models that are conceptually between the \ac{VAE} and \ac{UAE}. Additional models employing different combinations of the loss function components are provided in Appendix~\ref{sec:ablation}, Tab.~\ref{tab:all_loss_functions}.

\begin{table*}
	\centering
	\caption{A comparison of the \ac{VAE}, \ac{RAE}-GP (employing a Gradient Penalty (GP) on the decoder, a less general version of Eq. (\ref{eq:uae_dec_reg})), and \ac{UAE} loss functions, including the intermediate models \ac{UT}-\ac{VAE}, \ac{VAE}*, \ac{UT}-\ac{VAE}*, (weights omitted for clarity). \ac{UT}-\ac{VAE} uses the unscented transform in the \ac{VAE}, \ac{VAE}* uses the Wasserstein metric from Eq.~(\ref{eq:frob_approx}), and UT-\ac{VAE}* differs from the \ac{UAE} only in the lack of a decoder regularization term. All models use a diagonal posterior representation (except \ac{RAE}, which does not model uncertainty). The terms $\rvz$, $\vmu_\phi$, and $\bm{\sigma}_\phi$ are realized given the sample~$\rvx$.}
	\begin{tabular}{p{0.08\textwidth}p{0.6\textwidth}p{0.25\textwidth}}
		& Loss function & Posterior sampling \\
		\toprule
		$\mathcal{L}_{\text{\ac{VAE}}}$ & ${\scriptstyle\frac{1}{K}\sum^K_{k=1}  \Vert \rvx - D_\theta(\rvz_k)\Vert^2_2 +  \Vert\vmu_\phi\Vert^2_2 - n + \sum_i\sigma^2_{\phi, i} - 2\log \sigma_{\phi, i} }$ & ${\scriptstyle \rvz_k = \vmu_\phi + \bm{\sigma}_\phi \odot \bm{\epsilon}_k, \hspace{.1cm} \bm{\epsilon}_k\sim\mathcal{N}(\mathbf{0}, \mathbf{I})}$ \\
		\midrule
		$\mathcal{L}_{\text{UT-\ac{VAE}}}$ & ${\scriptstyle \Vert \rvx - \frac{1}{K}\sum^K_{k=1}D_\theta(\rvz_k)\Vert^2_2 +  \Vert\vmu_\phi\Vert^2_2 - n + \sum_i\sigma^2_{\phi, i} - 2\log \sigma_{\phi, i} }$ & ${\scriptstyle \rvz_k\sim \{\vchi_i(\vmu_\phi, \text{diag}(\bm{\sigma}^2_\phi))\}^{2n}_{i=0} }$ \\
		\midrule
		$\mathcal{L}_{\text{\ac{RAE}-GP}}$ & ${\scriptstyle \Vert \rvx - D_\theta(\rvz)\Vert^2_2 + \Vert \rvz \Vert^2_2 + \Vert\nabla_\rvz D_\theta(\rvz)\Vert^2_2 }$ & ${\small{\text{None, }}\scriptstyle \hspace{.1cm}\rvz=\vmu_\phi}$ \\
		\midrule
		$\mathcal{L}_{\text{\ac{VAE}*}}$ & ${\scriptstyle \frac{1}{K}\sum^K_{k=1}\Vert \rvx - D_\theta(\rvz_k)\Vert^2_2 +  \Vert\vmu_\phi\Vert^2_2 + \Vert \text{diag}(\bm{\sigma}^2_\phi) - \mathbf{I} \Vert^2_F}$ & ${\scriptstyle \rvz_k = \vmu_\phi + \bm{\sigma}_\phi \odot \bm{\epsilon}_k, \hspace{.1cm} \bm{\epsilon}_k\sim\mathcal{N}(\mathbf{0}, \mathbf{I})}$ \\
		\midrule
		$\mathcal{L}_{\text{UT-\ac{VAE}*}}$ & ${\scriptstyle \Vert \rvx - \frac{1}{K}\sum^K_{k=1}D_\theta(\rvz_k)\Vert^2_2 +  \Vert\vmu_\phi\Vert^2_2 + \Vert \text{diag}(\bm{\sigma}^2_\phi) - \mathbf{I} \Vert^2_F }$ & ${\scriptstyle \rvz_k\sim \{\vchi_i(\vmu_\phi, \text{diag}(\bm{\sigma}^2_\phi))\}^{2n}_{i=0} }$ \\
		\midrule
		$\mathcal{L}_{\text{\ac{UAE}}}$ & ${\scriptstyle \Vert \rvx - \frac{1}{K}\sum^K_{k=1}D_\theta(\rvz_k)\Vert^2_2 + \Vert\vmu_\phi\Vert^2_2 + \Vert \text{diag}(\bm{\sigma}^2_\phi) - \mathbf{I} \Vert^2_F + {\max}(\bm{\sigma}^2_\phi) \Vert \nabla_{\vmu_\phi} D_\theta(\vmu_\phi)\Vert^2_2 }$ & ${\scriptstyle \rvz_k\sim \{\vchi_i(\vmu_\phi, \text{diag}(\bm{\sigma}^2_\phi))\}^{2n}_{i=0} }$  \\	
		\bottomrule
	\end{tabular}
	\label{tab:loss_functions}
\end{table*}

\subsection{Sampling From the Prior-Less \ac{UAE}}\label{subsec:sampling}
Since the \ac{UAE} model doesn't regularize the aggregated posterior toward the prior using the \ac{KL} divergence~\citep{hoffman2016elbo} or the Wasserstein metric~\citep{patrini2020sinkhorn} (we use the per-posterior Wasserstein metric), it is not equipped with an easy-to-use sampling procedure as the \ac{VAE}. To remedy this, we use the straightforward ex-post density estimation procedure described in~\cite{ghosh2019variational} for the deterministic \ac{RAE} model. We fit the latent means $\vmu_\phi$ for each input sample $\rvx$ to a 10-component \ac{GMM} (which has shown good performance and generalization ability in the experiments of~\cite{ghosh2019variational} even for \ac{VAE} models) and use the mixture to sample from the latent space. For a fair comparison, we utilize this procedure in all models. 
\subsection{\ac{ELBO} Derivation}\label{subsec:elbo_derivation}
In the following, we analytically derive the \ac{UAE} model in Eq.~(\ref{eq:uae_obj}). The derivation is largely inspired from~\cite{ghosh2019variational}, with a few crucial differences allowing for greater generalizability and less restrictive assumptions. We start with the general \ac{ELBO} minimization formulation in Eq.~(\ref{eq:elbo}), augmented with a constraint
\begin{align}\label{eq:elbo_in_derivation}
    \argmin&_{\phi, \theta} \quad E_{x\sim p_\text{data}} \mathcal{L}_{\text{REC}} + \mathcal{L}_{\text{KL}}\\
 \begin{split}
	&\textrm{s.t.} \quad \Vert D_\theta(\rvz_1) - D_\theta(\rvz_2) \Vert_p < \epsilon,\ \\ &\rvz_1, \rvz_2 \sim q_\phi(\rvz|\rvx),\ \forall \rvx\sim p_\text{data} .
	\end{split} \label{eq:constraint}
\end{align}
Here, the decoder outputs given any two latent vectors $\rvz_1$ and $\rvz_2$ (any two draws from the posterior $q_\phi(\rvz|\rvx)$) are bounded via their $p$-norm difference, for a deterministic decoder $D_\theta$.
It was shown in \cite{ghosh2019variational} that the constraint in Eq.~(\ref{eq:constraint}) can be reformulated as
\begin{equation}
	\sup \{\Vert \nabla_\rvz D_\theta(\rvz)\Vert_p \} \cdot \sup \{\Vert \rvz_1 - \rvz_2 \Vert_p \} < \epsilon \ . \label{eq:predev}
\end{equation} 
We provide the full derivation in Appendix~\ref{sec:elbo_constraint}. In Eq.~(\ref{eq:predev}), $\nabla_\rvz D_\theta(\rvz)$ is the derivative of the decoder output w.r.t. its input (not the parameterization~$\theta$). The second term in the product depends on the parameterization of the posterior $q_\phi(\rvz|\rvx)$. For a Gaussian, $\sup \{\Vert \rvz_1 - \rvz_2 \Vert_p \}$ becomes a functional $r$ of the posterior entropy, $r(\mathbb{H}(q_\phi(\rvz|\rvx)))$. At this point, the \ac{RAE} derivation from~\cite{ghosh2019variational} takes a strong simplifying assumption of constant entropy for all samples $\rvx$, effectively asserting constant variance in the posterior. This allows to incorporate a simplified version of Eq.~(\ref{eq:predev}) into Eq.~(\ref{eq:elbo_in_derivation}) via the Lagrange multiplier $\gamma$, obtaining the following \ac{RAE} loss function\footnote{In~\cite{ghosh2019variational}, the decoder gradient penalty from Eq.~(\ref{eq:rae_loss}) is the analytically derived regularization; alternatives such as weight decay and spectral norm are offered as well and can also be used in the \ac{UAE}.}
\begin{equation}\label{eq:rae_loss}
\mathcal{L}_{\text{RAE}} = \Vert \rvx - D_\theta(\rvz)\Vert^2_2 + \beta\Vert \rvz \Vert^2_2 + \gamma\Vert\nabla_\rvz D_\theta(\rvz)\Vert^2_2\ .
\end{equation}
Here, the KL-term from Eq.~(\ref{eq:elbo_in_derivation}) is approximated by $\Vert \rvz \Vert^2_2$ due to the constant variance assumption.

In the \ac{UAE} formulation, the samples $\rvz_1$ and $\rvz_2$ in Eq.~(\ref{eq:predev}) simply correspond to the sigma points of $q_\phi(\rvz|\rvx)$ parameterized by $E_\phi(\rvx) = \{\vmu_\phi(\rvx), \mSigma_\phi(\rvx)\}$. Therefore, the term $\sup \{\Vert \rvz_1 - \rvz_2 \Vert_p \}$ can be computed analytically as the largest eigenvalue $\lambda_{\max}$ of the covariance matrix $\mSigma_\phi$. We regularize the decoder in an \ac{RAE}-manner around the posterior mean with $\Vert \nabla_{\vmu_\phi} D_\theta(\vmu_\phi)\Vert_p$ to enforce smoothness. Finally, the \ac{UAE} does not require the constant variance assumption; we can incorporate a posterior KL-term or the Wasserstein metric used in Eq.~(\ref{eq:uae_obj}). Thus, we arrive at the following analytical \ac{UAE} loss function from Eq.~(\ref{eq:uae_obj}) 
\begin{align}\label{eq:uae_obj_full}
	\begin{split}
		\mathcal{L}_{\text{UAE}} &= E_{\rvx\sim p_\text{data}} \mathcal{L}_{\text{REC}} + \beta\mathcal{L}_{\text{W}} + \\ +&\gamma\lambda_{\max}(\mSigma_\phi) \Vert \nabla_{\vmu_\phi} D_\theta(\vmu_\phi)\Vert_p  \ ,
	\end{split}
\end{align}
where a more general form of the Eq.~(\ref{eq:predev}) constraint is used than in Eq.~(\ref{eq:rae_loss}).

It follows from the derivation that the major difference between the \ac{RAE} on the one hand and \ac{VAE} and \ac{UAE} on the other is that the \ac{RAE} assumes constant variance in mapping the training data distribution into the latent space, thus not including any variance-compensating terms in the loss function. In effect, the \ac{RAE} considers all the dimensions equally and cannot take into account that the encoder might have different uncertainty per dimension and data point. Additionally, the difference between \ac{VAE} and \ac{UAE} is that the \ac{VAE} incorporates a sampling procedure with higher variance than the deterministic sigma-point sampling used in the unscented transform. Therefore, loss function-wise, the \ac{UAE} can be regarded as a middle-ground between the \ac{VAE} and \ac{RAE} -- deterministic and lower-variance in training than the \ac{VAE}, but with greater generalization capabilities than the \ac{RAE} due to the probabilistic formulation. 

\subsection{Posterior Regularization via the Wasserstein Metric}\label{subsec:kl_div_approx}
The usage of the Wasserstein metric is motivated by practical properties of \ac{VAE} model optimization. The training can be sensitive to the weighting of the \ac{KL} divergence term, which can lead to posterior collapse~\citep{dai2020usual}. The main factor is the strong variance regularization of the \ac{KL} divergence with its $\log$ term, which can be written as
\begin{align}\label{eq:kl_div_log}
\begin{split}
\mathcal{L}_{\text{KL}} = \Vert\vmu_\phi\Vert^2_2 + \text{tr}(\mSigma_\phi) - n - 2 \textstyle\sum_i\log {L}_{\phi, ii} 
\end{split}
\end{align}
If the posterior gets more peaked, which might be necessary for good reconstructions, the divergence quickly grows toward infinity. We observed such problems in particular with full-covariance posteriors (see Appendix~\ref{sec:ablation_full_cov}). 

Despite these problems the \ac{KL} divergence is theoretically sound. It was shown in \cite{hoffman2016elbo} that $\KL(q_\phi (\rvz|\rvx) \Vert p(\rvz))$ can be reformulated into two terms, one that weakly pushes toward overlapping per-sample posterior distributions and a \ac{KL} divergence between the aggregated posterior and the prior. The latter is required if samples are drawn from the prior and the former prevents the latent encoding from becoming a lookup table~\citep{mathieu2019disentangling}. Replacing the \ac{KL} divergence with the Wasserstein-2 metric preserves the tendency toward overlapping posteriors, but does not match the aggregated posterior to a predefined prior. However, a simple connection can be found to such models, see Appendix~\ref{sec:connection_wae}. Nevertheless, this matching is not required in our setup due to the ex-post density estimation. Furthermore, successful practical approaches like Stable Diffusion~\citep{rombach2022high} only require correctly learning the manifold and therefore do not need a certain aggregated posterior to sample from. 

We use the Wasserstein-2 metric between two Gaussian distributions. Mathematically, it can be written as 
\begin{align}\label{eq:wasserstein_analytical}
\begin{split}
W_2 (\mathcal{N}_1, \mathcal{N}_2) &= \Vert\vmu_\phi\Vert^2_2 + \text{tr}(\mSigma_\phi) + n - 2\text{tr}(\bm{\mSigma}_\phi^{1/2}) \\
&= \Vert\vmu_\phi\Vert^2_2 + \text{tr}(\mSigma_\phi) + n - 2\text{tr}(\bm{L}_\phi) \ ,
\end{split}
\end{align}
for $\mathcal{N}_1 = \mathcal{N}(\vmu_{\phi}, \mSigma_\phi)$ and $\mathcal{N}_2 = \mathcal{N}(\mathbf{0},\bm{I})$. The last three terms can be reformulated into Eq.~(\ref{eq:frob_approx})
\begin{align}\label{eq:kl_div_approx_deriv}
\begin{split}
&\text{tr}(\mSigma_\phi) + n - 2\text{tr}(\bm{L}_\phi) = \text{tr}(\bm{L}^T_\phi\bm{L}_\phi - 2\bm{L}_\phi+\bm{I}) = \\ &= \text{tr}((\bm{L}_\phi - \mathbf{I})^T(\bm{L}_\phi - \mathbf{I})) = \Vert \bm{L}_\phi - \mathbf{I} \Vert^2_F \ .
\end{split}
\end{align}
Disregarding the constant terms, it is clear that Eq. (\ref{eq:kl_div_log}) and Eq. (\ref{eq:wasserstein_analytical}) differ in the lack of the $\log$ term that infinitely penalizes zero-variance latents. In contrast, the Wasserstein metric even allows the posterior variance to approach zero if it helps to significantly reduce the reconstruction loss. This is evidenced in the aggregated posterior visualization of our model provided in Appendix~\ref{sec:agg_post}. 

Naturally, the reduced reconstruction losses brought on by the \textit{per-sample} Wasserstein metric in place of the KL divergence come at the cost of losing the \ac{ELBO} formulation of the overall optimization problem. Furthermore, the Wasserstein distance between the \textit{aggregated} posterior and the standard normal prior \citep{patrini2020sinkhorn} is not optimized either. Nevertheless, our empirical analysis shows that replacing the \ac{KL} divergence with a Wasserstein metric regularization of the per-sample posterior results in significantly better reconstruction performance.




%% file: chapters/results.tex
\section{Results}\label{sec:results}
In the following, we present quantitative and qualitative results of the \ac{UAE} and its precursors compared to the \ac{VAE} and \ac{RAE} baselines on Fashion-MNIST~\cite{xiao2017fashion}, CIFAR10~\citep{krizhevsky2009learning}, and CelebA~\citep{liu2015faceattributes}. We aim to delineate the effects of the \ac{UT} (along with the reconstruction loss in~Eq.~(\ref{eq:rec_loss}), Wasserstein metric, and the decoder regularization. Furthermore, we investigate multi-sampling and various sigma-point heuristics in Appendix~\ref{sec:multi_sigma} and ablate the entire loss function from Eq.~(\ref{eq:uae_obj}) in Appendix~\ref{sec:ablation}. In addition to evaluating the reconstruction and sampling quality (using a mixture for all models, see Sec. \ref{subsec:sampling}), we investigate if sampling only at the sigmas in training preserves the latent space structure (e.g. does not create 'holes') by evaluating interpolated samples. The metric is the widely-used \ac{FID}~\citep{heusel2017gans}, which quantifies the distance between two distributions of images. Detailed information about the network architecture, training, and the choice of \ac{FID} datasets is given in Appendix~\ref{sec:network}. 

\renewcommand\arraystretch{0.85}
\begin{table*}[]
	\centering
	\caption{Comparison of the architectures from Tab. \ref{tab:loss_functions}. In all sampling instances, we select 8 random samples or sigma points. In the unscented transform models (UT-VAE, UT-VAE*, UAE), we select random sigma points on all datasets apart from CIFAR10, where pairs of sigma points along the largest eigenvalue axes are selected (see Appendix~\ref{sec:multi_sigma}). All \ac{RAE} variants from~\cite{ghosh2019variational} are provided: \ac{RAE}-$\text{no-reg.}$ without decoder regularization, \ac{RAE}-GP with the Gradient Penalty (GP) from Eq.~(\ref{eq:rae_loss}), \ac{RAE}-L2 with decoder weight decay, and \ac{RAE}-SN with spectral normalization.}
	\begin{tabularx}{1\textwidth}{l *{10}{Y}}
		\toprule
		& \multicolumn{3}{c}{Fashion-MNIST}  
		& \multicolumn{3}{c}{CIFAR10}  
		& \multicolumn{3}{c}{CelebA}\\
		\cmidrule(lr){2-4} \cmidrule(lr){5-7} \cmidrule(lr){8-10} 
		& Rec. & Sample & Interp. & Rec. & Sample & Interp. & Rec. & Sample & Interp.\\
		\midrule
		\ac{VAE}$_{8\text{x}}$ & 44.29 & 48.73 & 61.99 & 110.0 & 120.6	& 118.3 & 65.86 & 68.53 & 68.75  \\
		UT-VAE$_{8\text{x}}$ & 27.79 & \textbf{30.39} & \textbf{39.92} & 91.04 & 111.7 & 104.3 & 50.11 & 54.15 & 54.32 \\
		\midrule
		\ac{RAE}-$\text{no-reg.}$ & 21.56 &	34.79 &	50.27 & 86.79 & 102.1 & 96.80 & 40.79 & 47.88 & 49.97 \\
		\ac{RAE}-GP & 22.91 & 33.80 & 50.74 & 85.70 & 100.7 & 96.06 & 39.89 & 46.67 & 46.18 \\
		\ac{RAE}-L2 & \textbf{20.28} & 32.06 & 48.52 & 84.27 & 99.26 & 94.23 & 38.78 & 46.44 & 50.33 \\	
		\ac{RAE}-SN & 21.40 & 33.50 & 49.60 & 85.75 & 101.1 & 96.48 & 41.23 & 48.39 & 50.23 \\
		\midrule
		\ac{VAE}*$_{8\text{x}}$ & 27.36 & 36.63 & 52.61 & 82.22 & 99.11 & 92.84 & 45.02 & 50.81 & 53.64 \\
		UT-VAE*$_{8\text{x}}$ & 23.64 & 31.51 & 48.06 & 81.12 & 100.6 & 93.80 & 40.18 & 47.39 & 49.62 \\
		\ac{UAE}$_{8\text{x}}$ & 25.07 & 35.19 & 54.24 & \textbf{71.97} & \textbf{89.91} & \textbf{83.50} & \textbf{38.48} & \textbf{45.60} & \textbf{45.88} \\
		\bottomrule
	\end{tabularx}  
	\label{tab:main_results}
\end{table*}

\renewcommand\arraystretch{0.85}
\begin{table*}[]
	\centering
	\caption{Comparison of a VAE model using the reconstruction loss of the mean image of random samples from the posterior: $\Vert \rvx - \frac{1}{K}\sum^K_{k=1} D_\theta(\rvz_k)\Vert^2_2, \hspace{.1cm} \rvz_k = \vmu_\phi + \bm{\sigma}_\phi \odot \bm{\epsilon}_k, \hspace{.1cm} \bm{\epsilon}_k\sim\mathcal{N}(\mathbf{0}, \mathbf{I})$, denoted by \ac{VAE}$^\dagger_{2\text{x}}$, and a model with the mean reconstruction loss of sigma points from the posterior: $\frac{1}{K}\sum^K_{k=1} \Vert \rvx - D_\theta(\rvz_k)\Vert^2_2, \hspace{.1cm} \rvz_k\sim \{\vchi_i(\vmu_\phi, \text{diag}(\bm{\sigma}^2_\phi))\}^{2n}_{i=0}$, denoted by UT-VAE$^\ddagger_{2\text{x}}$. The UT-VAE$_{2\text{x}}$ uses the full unscented transform with the reconstruction loss of the mean image of sigma points from the posterior: $\Vert \rvx - \frac{1}{K}\sum^K_{k=1} D_\theta(\rvz_k)\Vert^2_2, \hspace{.1cm} \rvz_k\sim \{\vchi_i(\vmu_\phi, \text{diag}(\bm{\sigma}^2_\phi))\}^{2n}_{i=0}$, as consistent with the Unscented Transform in Eq.~(\ref{eq:sigma}-\ref{eq:sigma_mu_transformation}). In the sigma-point variants of UT-VAE$^\ddagger_{2\text{x}}$ and UT-VAE$_{2\text{x}}$, random sigma points are selected for Fashion-MNIST and CelebA, while largest-eigenvalue pairs are used in CIFAR10.}
	\begin{tabularx}{1\textwidth}{l *{10}{Y}}
		\toprule
		& \multicolumn{3}{c}{Fashion-MNIST}  
		& \multicolumn{3}{c}{CIFAR10}  
		& \multicolumn{3}{c}{CelebA}\\
		\cmidrule(lr){2-4} \cmidrule(lr){5-7} \cmidrule(lr){8-10} 
		& Rec. & Sample & Interp. & Rec. & Sample & Interp. & Rec. & Sample & Interp.\\
		\midrule
		\ac{VAE}$_{2\text{x}}$ & 43.66 & 49.01 & 61.03 & 112.7 & 123.2 & 120.6 & 67.29 & 69.92 & 70.00 \\
		\ac{VAE}$^\dagger_{2\text{x}}$& 42.22 & 47.33 &	59.47 & 110.0 & 121.6 &	118.6 & 61.71 &	65.77 &	65.29 \\
		UT-VAE$^\ddagger_{2\text{x}}$& 46.79 & 52.87 & 74.11 & 115.2 & 128.2 & 124.7 & 54.61 & 61.03 & 59.49 \\
		UT-VAE$_{2\text{x}}$ & \textbf{36.25} & \textbf{40.30} & \textbf{53.10} & \textbf{95.70} & \textbf{115.4} & \textbf{107.3} & \textbf{51.61} & \textbf{57.42} & \textbf{56.56} \\	
		\bottomrule
	\end{tabularx}
	\label{tab:vae_mean_image}
\end{table*}
\renewcommand\arraystretch{1}

\renewcommand\arraystretch{0.2}
\begin{figure*}[]
	\centering
	\resizebox{1.7\columnwidth}{!}{%
		\begin{tabularx}{\textwidth}{tmmm}
			& \small Reconstruction & \small Sampling & \small Interpolation \\
			\toprule
			GT & \includegraphics[width=0.35\linewidth]{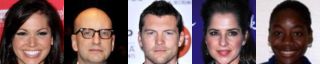} & & \\
			\ac{VAE} & \includegraphics[width=0.35\linewidth]{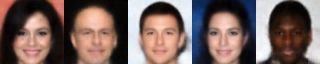} & \includegraphics[width=0.35\linewidth]{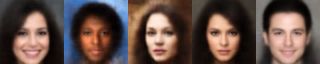} & \includegraphics[width=0.35\linewidth]{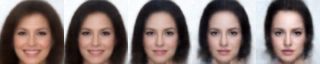}\\
			\ac{RAE} & \includegraphics[width=0.35\linewidth]{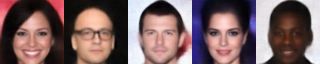} & \includegraphics[width=0.35\linewidth]{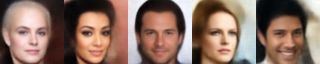} & \includegraphics[width=0.35\linewidth]{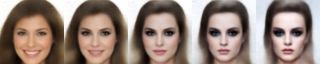} \\
			\ac{UAE} & \includegraphics[width=0.35\linewidth]{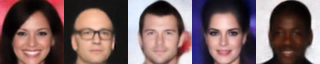} & \includegraphics[width=0.35\linewidth]{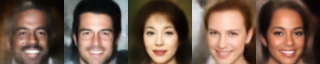} & \includegraphics[width=0.35\linewidth]{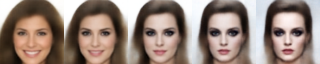} \\
			\bottomrule
	\end{tabularx}}
	\caption{Qualitative results on the CelebA dataset of the \ac{VAE}$_{8\text{x}}$, \ac{RAE}-L2, and \ac{UAE}$_{8\text{x}}$ models.}
	\label{fig:qualitative_results_celeba}
\end{figure*}
\renewcommand\arraystretch{1}

The main results are provided in Tab.~\ref{tab:main_results}. The table is divided into three parts: the first part shows the effects of applying the Unscented Transform to the vanilla \ac{VAE} model; the second part shows the baseline results of the~\ac{RAE}, while the third part shows the results of Wasserstein metric models. In the UT-VAE row of Tab.~\ref{tab:main_results}, \textit{we tweak the \ac{VAE} sampling to select instances at the sigma points while averaging the resulting images in the reconstruction loss, as consistent with the definition in Eq.~(\ref{eq:sigma}-\ref{eq:sigma_mu_transformation}). This simple change brings a remarkable near 40\% improvement on Fashion-MNIST on average, near 15\% on CIFAR10, and near 30\% on CelebA}. It provides strong evidence that a higher-quality, lower-variance representation of the posterior distribution results in higher-quality decoded images.

The deterministic baseline \ac{RAE} model in Tab.~\ref{tab:main_results} sets the context with a significantly higher performance than the vanilla \ac{VAE}. The Wasserstein metric of the \ac{VAE}*, which preserves the latent space regularization in spirit of the \ac{RAE} but extends it to a probabilistic, non-constant variance setting, can be considered close to the non-regularized \ac{RAE}: outperforms it on CIFAR10 while being behind on Fashion-MNIST and CelebA. More importantly, the VAE* model also achieves a large improvement over the classical \ac{VAE} in all metrics and on all datasets, achieved effectively only by replacing the logarithm term with a linear term. This indicates that the rigidity of the KL divergence w.r.t. posterior variance potentially harms the quality of decoded samples, particularly on the richer CIFAR10 and CelebA.

Observing the UT-\ac{VAE}* row in Tab.~\ref{tab:main_results}, it can be seen that the unscented transform (UT) sampling in the \ac{VAE}* context gives a further, albeit lesser boost in most metrics than with the \ac{KL} divergence. Due to the Wasserstein metric's ability to shrink the posterior variance while approaching convergence, the effect of any sampling is reduced. Nevertheless, it provides a considerable, approximately 10\% boost on CelebA and Fashion-MNIST as well as a larger relative improvement with multiple samples than in VAE* (see Tab.~\ref{tab:ut_heuristics_mean_sigma}, \ref{tab:multi-sample} in Appendix~\ref{sec:multi_sigma}). Finally, the generalized decoder regularization from Eq.~(\ref{eq:uae_dec_reg}) of the \ac{UAE} applies a strong smoothing effect and further boosts the performance on CelebA and especially CIFAR10. Surprisingly, it yields a regression on Fashion-MNIST; similar effect of the gradient penalty harming the \ac{RAE} performance compared to no-regularization is observable in~\cite{ghosh2019variational} MNIST experiments. Overall, compared to the \ac{RAE}, the \ac{UAE} achieves significant improvements on CIFAR10 and a minor improvement on CelebA, while interestingly, the best model on Fashion-MNIST can be considered the UT-VAE. 

In Tab.~\ref{tab:vae_mean_image}, we take a deeper look at the performance of the \ac{UT} reconstruction loss term from Eq.~(\ref{eq:rec_loss}). We empirically compare two strategies for designing the loss function:~(i) use the mean reconstruction loss of images for each selected sample from the posterior (consistent with the standard VAE reconstruction loss) and~(ii) apply the reconstruction loss to the mean image of samples from the posterior. Quantitative results in Tab.~\ref{tab:vae_mean_image} consistently show the advantages of strategy~(ii) for both the VAE and UT-VAE models using random samples and sigma points, respectively.

CelebA qualitative results are shown in Fig.~\ref{fig:qualitative_results_celeba} and reflect the \ac{FID} scores: the \ac{UAE} images appear similar to the \ac{RAE} but significantly more realistic than the \ac{VAE}. Fashion-MNIST and CIFAR10 images are provided in Appendix~\ref{sec:qualitative}.

%% file: chapters/conclusion.tex
\section{Conclusion}
In this paper, we introduced a novel \ac{VAE} architecture employing the Unscented Transform, a lower-variance alternative to the reparameterization trick. We have challenged one of the core components of the \ac{VAE} by showing that a sigma-point transform of the posterior significantly outperforms propagating random samples through the decoder. This was empirically shown for a small number of sigma points ($2$, $4$, and $8$) while taking more becomes impractical due to computationally-intensive training. Additionally, we proposed to use the Wasserstein metric, which does not optimize the \ac{ELBO}. Although it can be considered as the main theoretical limitation of our model, it is a sound practical alternative to the \ac{KL} divergence. By breaking its rigidity w.r.t. posterior variance, we unlocked performance improvements brought on by sharper posteriors that preserve a smooth latent space. Our work contributes an important step toward establishing competitive deterministic and deterministic-sampling generative models. Future work will thus focus on expanding the classes of supported generative models and on evaluation of further deterministic and quasi-deterministic sampling methods.

%% file: chapters/appendix.tex
\section{Network Architecture and Training}\label{sec:network}
\begin{table}[h!]
	\scriptsize{
		\caption{Network architectures of the implemented \ac{VAE}, \ac{RAE}, and \ac{UAE} models. Batch dimensions omitted for clarity.}\label{tab:networks}
		\begin{tabularx}{\textwidth}{X}
			\\
			\toprule
			\ac{VAE}, \ac{UAE}:\hspace{0.1cm} $\rvx_{C\times W\times H} \rightarrow $ \textsc{Encoder} $\rightarrow$ \{$\textsc{FC}_{1024\times n}: \vmu_\phi, \textsc{FC}_{1024\times n}: \log\bm{\sigma}_\phi^2\} \rightarrow \rvz $ $\rightarrow$ \textsc{Decoder} $\rightarrow$ $\hat{\rvx}$ \\
			\midrule
			\ac{RAE}:\hspace{0.1cm} $\rvx_{C\times W\times H} \rightarrow $ \textsc{Encoder} $\rightarrow$ \{$\textsc{FC}_{1024\times n}: \rvz_\phi\} $ $\rightarrow$ \textsc{Decoder} $\rightarrow$ $\hat{\rvx}$\\
			\midrule
			\textsc{Encoder}: $\textsc{Conv}_{32\times64}\rightarrow\textsc{Conv}_{64\times128}\rightarrow\textsc{Conv}_{128\times256}\rightarrow\textsc{Conv}_{256\times512}\rightarrow\textsc{Conv}_{512\times1024}\rightarrow\textsc{Flatten}$ \\
			\midrule
			\textsc{Decoder}: $\textsc{FC}_{n\times1024\cdot8\cdot8}\rightarrow\textsc{TConv}_{1024\times512}\rightarrow\textsc{TConv}_{512\times256}[\rightarrow\textsc{TConv}_{256\times128}]^{\text{CelebA}}\rightarrow\textsc{TConv}_{256\hspace{0.03cm}\text{or}\hspace{0.03cm}128\times C} $\\
			\midrule
			\textsc{MNIST}: $C=1, W=H=32, n=64$\\
			\textsc{CIFAR10}: $C=3, W=H=32, n=128$\\
			\textsc{CelebA}: $C=3, W=H=64, n=64$\\
			\bottomrule
		\end{tabularx}
	}	
\end{table}

Network architectures are given in Tab. \ref{tab:networks} and largely follow the architecture in~\cite{ghosh2019variational}. For consistency, all models share the same encoder/decoder structure. All encoder 2D convolution blocks contain $3\times3$ kernels, stride $2$, and padding $1$, followed by a 2D batch normalization and a Leaky-ReLU activation. The decoder transposed convolutions share the same parameters as the encoder convolutions apart from using a $4\times4$ kernel. The last transposed convolution (mapping to channel dimension) however has a $3\times3$ kernel and is followed by a $\tanh$ activation (without batch normalization).

The dataset preprocessing procedure is the following. The Fashion-MNIST images are scaled from $28\times28$ to $32\times32$. For the training dataset, we use 50k out of the 60k provided examples, leaving the remaining 10k for the validation dataset. For the test dataset, we use the provided examples. In CIFAR10, we perform a random horizontal flip on the training data followed by a normalization for all dataset subsets. We use the same training/validation/test split method as in Fashion-MNIST. In CelebA, we perform a $148\times148$ center crop and resize the images to $64\times64$. We use the provided training/validation/testing subsets.

All models are implemented in PyTorch~\citep{paszke2019pytorch} and use the library provided in~\cite{Seitzer2020FID} for \ac{FID} computation. The models are trained for 100 epochs, starting with a $0.005$ learning rate that is then halved after every five epochs without improvement. The weights used in the loss functions are the following: KL-divergence (or the Wasserstein metric) terms are weighted with $\beta=2.5e^{-4}$ in the case of \ac{VAE} and \ac{UAE} and $\beta=1e^{-4}$ for the \ac{RAE}. The decoder regularization terms are weighted with $\gamma=1e^{-6}$ for both \ac{RAE} and \ac{UAE}. We performed minimal hyperparameter search over the weights. 

In computing the \ac{FID} scores, we follow the same procedure as in~\cite{ghosh2019variational}. In the three cases of reconstruction, sampling, and interpolation, we evaluate the \ac{FID} to the test set image reconstructions as the ground-truth. In the reconstruction metric, we use the validation set image reconstructions. In sampling, we fit the training dataset latent features to a \ac{GMM} (see Sec.~\ref{subsec:sampling}) and sample and reconstruct the same number of elements as in the test set. In interpolation, we apply mid-point spherical interpolation between a random pair of validation set embeddings. In all cases, we generate a single image per input; this image corresponds to the posterior mean of the latent distribution. This mean latent feature vector is also used in sampling and interpolation while fitting a mixture ex-post or interpolating the latent space vectors. Thus, the resulting number of generated images for \ac{FID} computation is the same regardless of the number of sigma points or samples used in training. In all experiments, the average \ac{FID} score of three runs is reported, while observing a similar variation between scores of individual runs among the models employing the \ac{UT} compared to the vanilla \ac{VAE}. In contrast, the scores of \ac{RAE} and \ac{VAE}* modes were significantly more consistent.

The network architectures largely follow the structure adopted by~\cite{ghosh2019variational}, with the difference of the added first two encoder layers. Nevertheless, in Tab. \ref{tab:main_results}, we did not manage to reproduce the \ac{FID} values reported in~\cite{ghosh2019variational} on CelebA and CIFAR10, even observing that removing the first two encoder layers reduces the overall performance. We suspect that it is due to the differing Tensorflow and PyTorch model implementations as well as the \ac{FID} computation libraries. However, in most cases, our implementation of the \ac{RAE} attains a significantly larger performance gain over the \ac{VAE} than reported in~\cite{ghosh2019variational}.
\newpage

\section{Gradient Variance and Bias}\label{sec:grad_variance}
In this section, we investigate the gradient variance and bias of the proposed base \ac{UT}-VAE model. Compared to random sampling of the reparameterization trick, using a different integration scheme like sampling sigma points can be biased. It can nevertheless achieve lower variance depending on the nonlinear function of the decoder. Thus, for our decoder setup, we compare the gradient variance and bias of the \ac{UT}-\ac{VAE} (with random sigma pair sampling) and the \ac{VAE}$^\dagger$ (with random sampling) employing the decoder output mean instead of the sample mean\footnote{Reconstruction loss function of the \ac{VAE}$^\dagger$: $\Vert \rvx - \frac{1}{K}\sum^K_{k=1} D_\theta(\rvz_k)\Vert^2_2, \hspace{.1cm} \rvz_k = \vmu_\phi + \bm{\sigma}_\phi \odot \bm{\epsilon}_k, \hspace{.1cm} \bm{\epsilon}_k\sim\mathcal{N}(\mathbf{0}, \mathbf{I})$} (see Tab.~\ref{tab:vae_mean_image} for a performance comparison) in order to isolate the effect of sampling sigma points.

\begin{figure}[]
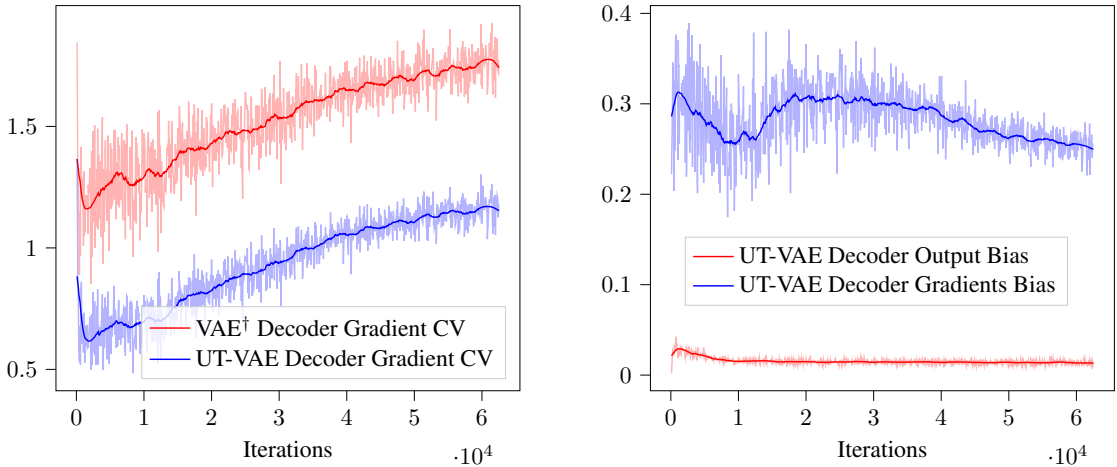

	\centering
	\begin{subfigure}[b]{0.45\columnwidth}
		\centering
		\scalebox{0.9}{\input{images/grad_var_coeff.tex}}
		\caption{Median of the decoder gradient \ac{CV} for \ac{UT}-\ac{VAE} and \ac{VAE}$^\dagger$}
		\label{fig:grad_var_coeff}
	\end{subfigure}\hspace{0.1cm}
	\begin{subfigure}[b]{0.45\columnwidth}
	\centering
	\scalebox{0.9}{\input{images/grad_decoder_bias.tex}}
		\caption{Median relative bias based on an estimate of the true gradient using 200 random samples}
		\label{fig:grad_bias}
	\end{subfigure}
	\caption{Comparison of the variance and bias trade-off for the \ac{VAE}$^\dagger$ (employing the decoder output mean instead of the sample mean, see Tab.~\ref{tab:vae_mean_image}) and \ac{UT}-\ac{VAE} across approx. 60k training steps (100 epochs) on the CIFAR10 dataset. The data is based on a single training of an \ac{UT}-\ac{VAE} where every 50th epoch the gradient variance and bias was estimated using different sampling schemes. In case of \ac{VAE}$^\dagger$, two random points are sampled (in accordance with the reparameterization trick), while in case of \ac{UT}-\ac{VAE}, a single sigma point pair is sampled.}
	\label{fig:grad_var_bias}
\end{figure}

We train both models and estimate the gradient variance and bias every 50th iteration. For \ac{UT}-\ac{VAE} we independently sample 50 sigma point pairs, pass them through the decoder, and calculate the gradients' mean $m_j$ and standard deviation $\sigma_j$. For \ac{VAE}$^\dagger$ we draw 2 random samples 200 times and perform the same steps to obtain $m'_j$ and $\sigma'_j$. We calculate the median \ac{CV} of the gradients for both models, assuming that $m'_j$ computed with 200 random samples is a good enough estimate of the true gradient. Furthermore, we compute the median relative bias $b_{rel}$ for the decoder gradients and output of the UT-VAE. The \ac{CV} (for UT-VAE) and $b_{rel}$ (for decoder gradients bias) are computed as follows
\begin{equation}
\text{CV} = \text{median} \left\lbrace \frac{\sigma_j}{|m_j|} \right\rbrace \qquad
b_{rel} = \text{median} \left\lbrace \frac{|m_j - m'_j|}{|m'_j|} \right\rbrace  . 
\end{equation}
The gradient variance results are depicted in Fig.~\ref{fig:grad_var_coeff}. The variance of the sigma pair sampling of the \ac{UT}-\ac{VAE} is consistently lower than the gradient variance of the random sampling within \ac{VAE}$^\dagger$. Interestingly, for the \ac{VAE}$^\dagger$ the standard deviation of the gradients is on average larger than the magnitude of the gradient during the whole training, whereas for the \ac{UT}-\ac{VAE} this is only the case at the end of the training. Fig.~\ref{fig:grad_bias} shows the relative decoder output bias as well as the relative gradient bias of the \ac{UT}-\ac{VAE} at the same iterations. Whereas the relative bias at the decoder output is below 3\% throughout the whole training, the bias of the gradients is around 30\% of their magnitude. It is unclear whether such a substantial gradient bias is behind the good performance of the \ac{UT}-\ac{VAE} or if there is a performance trade-off between variance and bias. Nevertheless, our experiments show that, under a common decoder architecture, integration schemes like the \ac{UT} can exhibit lower variance and higher bias while outperforming the standard \ac{VAE} sampling scheme. Thus, investigating alternative integration schemes for \ac{VAE}s can be a promising research direction.



\section{Additional Results: Multi-Sigma Heuristics and Multi-Sample Models}\label{sec:multi_sigma}
The UT-\ac{VAE} loss function defined in Tab. \ref{tab:loss_functions} samples $K$ sigma points in the reconstruction term. Increasing the number of sigma points (up to $2n+1$) improves the estimate of the transformed posterior distribution and thus the resulting reconstruction quality, at the expense of an approximately linear increase in training time. We observed this in most cases when training on $2$, $4$, and $8$ sigma points, see Tab.~\ref{tab:ut_heuristics_mean_sigma}. However, a much larger number of sigma points might not result in expected additional performance improvement due to significantly larger batch size, which could be mitigated by constructing approaches to select and train on a fixed, smaller batch size. 

For $K$ selected sigma points, various strategies can be used instead of sampling a discrete uniform distribution. For example, only pairs of sigma points along an axis can be chosen, conveying the width of the posterior distribution in the given dimension. This strategy can be adapted to select pairs along axes with largest eigenvalues. Tab.~\ref{tab:ut_heuristics_mean_sigma} also explores different sampling heuristics in the case of UT-VAE and UT-VAE*. We have observed that models trained with KL divergence exhibit larger variation in results w.r.t. the sampling heuristic, which is reasonable since the Wasserstein metric's posterior variance suppression diminishes the effect of sampling. The choice of the sigma-point selection heuristic turns out to have a large effect on the overall performance given a dataset. We have observed that a random selection of sigma points performs consistently well across all datasets while selecting random pairs generates reasonable results only in the case of CIFAR10. Interestingly, random-pairs performs very poorly on Fashion-MNIST and CelebA while largest eigenvalue pairs shows very good performance in the UT-VAE case on CIFAR10. In the main experiments of Tab.~\ref{tab:main_results}, we used a random selection for the Fashion-MNIST and CelebA models and largest-eigenvalue pairs for CIFAR10, due to its superior performance in the UT-VAE case.

Tab.~\ref{tab:multi-sample} analyzes models using multiple samples in training. We compare the VAE* and the \ac{UAE} with the classical \ac{VAE} and the \ac{IWAE}~\cite{burda2016iwae} as a baseline where multiple importance-weighted posterior samples help achieve a tighter lower bound. Observing the results, it is clear that models employing the Wasserstein metric can benefit from increasing the number of samples in training despite their ability to reduce the latent space variance, while significantly outperforming the baselines.



\begin{table}
	\centering
	\caption{Analysis of the number of sampled sigma points and different heuristics, where the mean image of multiple sigma points is matched to the ground truth in the reconstruction loss. The three investigated heuristics are sampling random sigma points, random pairs of sigma points along an axis, and pairs of sigma points along axes with largest eigenvalues.}
	\begin{tabularx}{1\textwidth}{l *{11}{Y}}
		\toprule
		& \multicolumn{3}{c}{Fashion-MNIST} & \multicolumn{3}{c}{CIFAR10} & \multicolumn{3}{c}{CelebA}\\
		\cmidrule(lr){2-4} \cmidrule(lr){5-7} \cmidrule(lr){8-10} 
		& Rec. & Samp. & Interp. & Rec. & Samp. & Interp. & Rec. & Samp. & Interp.\\
		\midrule
		UT-VAE$_{1\text{x}, \text{rand.}}$ & 47.27 & 52.10 & 67.16 & 119.9 & 129.8 & 127.9 & 55.93 & 62.13 &	60.54 \\
		UT-VAE$_{2\text{x}, \text{rand.}}$ & 36.25 & 40.30 & 53.10 & 111.5 & 124.7 & 121.0 & 51.61 & 57.42 & 56.56 \\
		UT-VAE$_{4\text{x}, \text{rand.}}$ & 32.13 & 36.41 & 47.30 & 105.9 & 119.8 & 115.9 & 50.85 & 55.82 & 55.99 \\
		UT-VAE$_{8\text{x}, \text{rand.}}$ & 27.79 & \textbf{30.39} & \textbf{39.92} & 95.40 & 110.8 & 106.4 & 50.11 & 54.15 & 44.32 \\
		UT-VAE*$_{2\text{x}, \text{rand.}}$ & 28.26 & 36.36 & 50.69 & 85.88 & 103.7 & 96.90 & 44.32 & 50.33 & 52.40 \\
		UT-VAE*$_{4\text{x}, \text{rand.}}$ & 24.38 & 32.75 & 49.40 & 81.99 & 100.6 & 93.52 & 42.52 &	49.21 &	51.35 \\
		UT-VAE*$_{8\text{x}, \text{rand.}}$ & \textbf{23.64} & 31.51 & 48.06 & 81.10 & 99.87 & 92.48 & \textbf{40.18} & \textbf{47.39} & \textbf{49.62} \\
		\midrule
		UT-VAE$_{2\text{x}, \text{rand. pairs}}$ & 102.1 & 115.1 & 112.8 & 102.3 &	119.6 & 114.0 & 150.0 & 150.4 &	151.3 \\
		UT-VAE$_{4\text{x}, \text{rand. pairs}}$ & 96.85 & 110.1 & 107.3 & 101.0 & 119.5 & 113.4 & 224.3 & 225.0 & 225.4 \\
		UT-VAE$_{8\text{x}, \text{rand. pairs}}$ & 90.14 & 103.6 & 101.5 & 100.3 &	119.2 &	113.2 & 173.2 & 175.4 &	175.8 \\
		UT-VAE*$_{2\text{x}, \text{rand. pairs}}$ & 32.66 & 38.68 &	58.72 & 85.64 & 102.3 &	97.00 & 45.96 &	53.16 &	51.49 \\
		UT-VAE*$_{4\text{x}, \text{rand. pairs}}$ & 32.85 &	38.58 &	57.70 & 84.62 & 102.2 &	96.14 & 252.9 &	254.8 &	253.8 \\
		UT-VAE*$_{8\text{x}, \text{rand. pairs}}$ & 30.65 &	36.88 &	56.42 & \textbf{80.51} & \textbf{98.40} & \textbf{91.96} & 141.9 & 144.3 & 147.4 \\
		\midrule
		UT-VAE$_{2\text{x}, \text{larg. } \lambda \text{ pairs}}$ & 106.6 &	118.6 &	115.7 & 95.70 &	115.4 &	107.3 & 54.02 & 60.29 & 60.26 \\
		UT-VAE$_{4\text{x}, \text{larg. } \lambda \text{ pairs}}$ & 108.3 &	120.1 &	117.2 & 92.56 & 111.6 & 104.2 & 46.37 & 53.53 & 52.62 \\
		UT-VAE$_{8\text{x}, \text{larg. } \lambda \text{ pairs}}$ & 115.5 &	128.8 &	126.3 & 91.04 & 111.7 & 104.3 & 48.59 & 55.22 & 55.29 \\
		UT-VAE*$_{2\text{x}, \text{larg. } \lambda \text{ pairs}}$ & 33.49 & 42.63 & 61.57 & 82.17 & 100.7 & 93.80 & 55.57 & 61.42 & 61.53 \\
		UT-VAE*$_{4\text{x}, \text{larg. } \lambda \text{ pairs}}$ & 34.94 & 43.18 & 67.65 & 81.61 & 101.3 & 94.11 & 48.41 & 54.70 & 54.80 \\
		UT-VAE*$_{8\text{x}, \text{larg. } \lambda \text{ pairs}}$ & 31.08 & 41.06 & 64.58 & 81.12 & 100.6 & 93.80 & 45.08 & 51.45 & 52.05 \\
		\bottomrule
	\end{tabularx}
	\label{tab:ut_heuristics_mean_sigma}
\end{table}

\begin{table}
	\centering
	\caption{Comparison of models employing multiple samples in training. The \ac{UAE} uses random sigma points on Fashion-MNIST and CelebA and largest-eigenvalue pairs on CIFAR10.}
	\begin{tabularx}{1\textwidth}{l *{11}{Y}}
		\toprule
		& \multicolumn{3}{c}{Fashion-MNIST} & \multicolumn{3}{c}{CIFAR10} & \multicolumn{3}{c}{CelebA}\\
		\cmidrule(lr){2-4} \cmidrule(lr){5-7} \cmidrule(lr){8-10} 
		& Rec. & Sample & Interp. & Rec. & Sample & Interp. & Rec. & Sample & Interp.\\
		\midrule
		\ac{VAE}$_{1\text{x}}$ & 45.64 & 49.99 & 61.33 & 116.4 & 126.8 & 124.2 & 68.32 & 71.05 & 71.16 \\
		\ac{VAE}$_{2\text{x}}$ & 43.66 & 49.01 & 61.03 & 112.7 & 123.2 & 120.6 & 67.29 & 69.92 & 70.00 \\
		\ac{VAE}$_{4\text{x}}$ & 44.94 & 49.51 & 62.29 & 111.7 & 121.3 & 119.5 & 66.32 & 68.87 & 69.06 \\
		\ac{VAE}$_{8\text{x}}$ & 44.29 & 48.73 & 61.99 & 110.0 & 120.6	& 118.3 & 65.86 & 68.53 & 68.75 \\
		\midrule
		\ac{IWAE}$_{1\text{x}}$ & 49.27 & 53.71 & 64.50 & 111.7 & 121.6 & 119.6 & 68.28 &	71.16 &	71.17 \\
		\ac{IWAE}$_{2\text{x}}$ & 48.21 & 53.11 & 65.69 & 112.1 & 122.4 & 119.8 & 66.85 &	69.81 &	69.74 \\
		\ac{IWAE}$_{4\text{x}}$ & 47.40 & 51.77 & 64.10 & 110.6 & 120.6 & 118.2 & 66.01 &	68.82 & 68.90 \\
		\ac{IWAE}$_{8\text{x}}$ & 46.16 & 50.91 & 63.68 & 108.9 & 118.9 & 116.9 & 64.83 &	67.96 &	67.86 \\
		\midrule
		\ac{VAE}*$_{1\text{x}}$ & 31.62 & 38.44 & 52.33 & 83.49 & 101.5 & 94.56 & 44.69 & 50.55 & 53.18 \\
		\ac{VAE}*$_{2\text{x}}$ & 30.07 & 37.92 & \textbf{52.15} & 84.57 & 102.2 & 95.61 & 45.18 & 50.97 & 53.73 \\
		\ac{VAE}*$_{4\text{x}}$ & 28.98 & 41.35 & 52.17 & 84.64 & 102.3 & 95.96 & 45.03 & 50.59 & 53.32 \\
		\ac{VAE}*$_{8\text{x}}$ & 27.36 & 36.63 & 52.61 & 82.22 & 99.11 & 92.84 & 45.02 & 50.81 & 53.64 \\
		\midrule
		\ac{UAE}$_{2\text{x}}$ & 29.29 & 37.59 & 53.69 & 77.71 & 96.37 & 89.71 & 40.07 & 47.28 & 50.51 \\
		\ac{UAE}$_{4\text{x}}$ & 27.11 & 38.03 & 53.11 & 75.63 & 93.02 & 86.41 & 39.48 & 46.35 & 50.94 \\
		\ac{UAE}$_{8\text{x}}$ & \textbf{25.07} & \textbf{35.19} & 54.24 & \textbf{71.97} & \textbf{89.91} & \textbf{83.50} & \textbf{38.48} & \textbf{45.60} & \textbf{45.88} \\
		\bottomrule
	\end{tabularx}
	\label{tab:multi-sample}
\end{table}


\section{Additional Results: Ablation Study of the Loss Components}\label{sec:ablation}

This section provides an additional ablation study of the loss components used in the \ac{UAE} model. The loss functions considered are provided in the upper half of Tab.~\ref{tab:all_loss_functions} and the obtained results are in Tab.~\ref{tab:ablation}. There are three dimensions along which the results can be interpreted: Wasserstein metric, unscented transform, and the generalized decoder regularization (gradient penalty).

Tab.~\ref{tab:ablation} is divided into two parts: the top part models use the analytical form of the KL divergence in Eq. (\ref{eq:kl_div_log_l}) while the bottom part use the Frobenius norm mismatch derived from the Wasserstein metric in Eq. (\ref{eq:frob_approx}). It is clearly visible that the latter models strongly outperform the former, in all datasets and configurations. The loss function allows for a sharper posterior and thus larger expressiveness of the model (see Appendix~\ref{sec:agg_post}).

Similarly, the unscented transform models UT-VAE and UT-VAE* clearly outperform the random sampling and per-sample reconstruction counterparts of VAE and VAE*. In the latter case, the differences are smaller due to the sharper posterior of the VAE*. An ablation study of the unscented transform components can be found in Tab.~\ref{tab:vae_mean_image}.

Considering the gradient penalty models, interesting interplays can be noticed. Applying the decoder regularization on the vanilla \ac{VAE} and the \ac{VAE}* (this model can be considered closest to the \ac{RAE}-GP) brings only minor improvements in the case of CIFAR10 and CelebA for each of the models respectively. The strong smoothing of the latent space however seems detrimental when combined with the unscented transform and the KL divergence training. One can conclude that only the latent space regularization models (such as the Wasserstein metric VAE* or the deterministic \ac{RAE}) can benefit from decoder regularization. Furthermore, the effect appears to be dataset-dependent since the Fashion-MNIST VAE* and UT-VAE* slightly regress when augmented with decoder regularization.

\begin{table}
	\caption{The loss functions used for the models in Tab.~\ref{tab:ablation} and Tab.~\ref{tab:full_covariance_ablation}. The upper and lower half of the table contain diagonal and full-covariance posterior models, respectively.}
	\resizebox{1\columnwidth}{!}{%
		\begin{tabularx}{\textwidth}{abs}
			& \small Loss function & \small Posterior sampling \\
			\toprule
			$\mathcal{L}_{\text{\ac{VAE}}}$ & ${\scriptstyle \frac{1}{K}\sum^K_{k=1}  \Vert \rvx - D_\theta(\rvz_k)\Vert^2_2 + \Vert\vmu_\phi\Vert^2_2 - n  + \sum_i\sigma^2_{\phi, i} - 2\log \sigma_{\phi, i} }$ & ${\scriptstyle \rvz_k = \vmu_\phi + \bm{\sigma}_\phi \odot \bm{\epsilon}_k, \hspace{.1cm} \bm{\epsilon}_k\sim\mathcal{N}(\mathbf{0}, \mathbf{I})}$ \\
			\toprule
			$\mathcal{L}_{\text{\ac{VAE}-GP}}$ & ${\scriptstyle \frac{1}{K}\sum^K_{k=1}  \Vert \rvx - D_\theta(\rvz_k)\Vert^2_2 + \Vert\vmu_\phi\Vert^2_2+ \sum_i\sigma^2_{\phi, i} - 2\log\sigma_{\phi, i} + {\max}(\bm{\sigma}_\phi) \Vert \nabla_{\vmu_\phi} D_\theta(\vmu_\phi)\Vert^2_2} $ & ${\scriptstyle \rvz_k = \vmu_\phi + \bm{\sigma}_\phi \odot \bm{\epsilon}_k, \hspace{.1cm} \bm{\epsilon}_k\sim\mathcal{N}(\mathbf{0}, \mathbf{I})}$ \\
			\midrule
			$\mathcal{L}_{\text{UT-\ac{VAE}}}$ & ${\scriptstyle \Vert \rvx - \frac{1}{K}\sum^K_{k=1} D_\theta(\rvz_k)\Vert^2_2 + \Vert\vmu_\phi\Vert^2_2 + \sum_i\sigma^2_{\phi, i} - 2\log \sigma_{\phi, i} }$ & ${\scriptstyle \rvz_k\sim \{\vchi_i(\vmu_\phi, \text{diag}(\bm{\sigma}^2_\phi))\}^{2n}_{i=0} }$ \\
			\midrule
			$\mathcal{L}_{\text{UT-\ac{VAE}-GP}}$ & ${\scriptstyle \Vert \rvx - \frac{1}{K}\sum^K_{k=1} D_\theta(\rvz_k)\Vert^2_2 + \Vert\vmu_\phi\Vert^2_2+ \sum_i\sigma^2_{\phi, i} - 2\log \sigma_{\phi, i} + {\max}(\bm{\sigma}_\phi) \Vert \nabla_{\vmu_\phi} D_\theta(\vmu_\phi)\Vert^2_2 }$ & ${\scriptstyle \rvz_k\sim \{\vchi_i(\vmu_\phi, \text{diag}(\bm{\sigma}^2_\phi))\}^{2n}_{i=0} }$ \\
			\midrule
			$\mathcal{L}_{\text{\ac{VAE}*}}$ & ${\scriptstyle \frac{1}{K}\sum^K_{k=1}  \Vert \rvx - D_\theta(\rvz_k)\Vert^2_2 + \Vert\vmu_\phi\Vert^2_2+ \Vert \text{diag}(\bm{\sigma}^2_\phi) - \mathbf{I} \Vert^2_F}$ & ${\scriptstyle \rvz_k = \vmu_\phi + \bm{\sigma}_\phi \odot \bm{\epsilon}_k, \hspace{.1cm} \bm{\epsilon}_k\sim\mathcal{N}(\mathbf{0}, \mathbf{I})}$ \\
			\midrule
			$\mathcal{L}_{\text{\ac{VAE}*-GP}}$ & ${\scriptstyle \frac{1}{K}\sum^K_{k=1}  \Vert \rvx - D_\theta(\rvz_k)\Vert^2_2 + \Vert\vmu_\phi\Vert^2_2+ \Vert \text{diag}(\bm{\sigma}^2_\phi) - \mathbf{I} \Vert^2_F + {\max}(\bm{\sigma}_\phi) \Vert \nabla_{\vmu_\phi} D_\theta(\vmu_\phi)\Vert^2_2}$ & ${\scriptstyle \rvz_k = \vmu_\phi + \bm{\sigma}_\phi \odot \bm{\epsilon}_k, \hspace{.1cm} \bm{\epsilon}_k\sim\mathcal{N}(\mathbf{0}, \mathbf{I})}$ \\
			\midrule
			$\mathcal{L}_{\text{UT-\ac{VAE}*}}$ & ${\scriptstyle \Vert \rvx - \frac{1}{K}\sum^K_{k=1} D_\theta(\rvz_k)\Vert^2_2 + \Vert\vmu_\phi\Vert^2_2+ \Vert \text{diag}(\bm{\sigma}^2_\phi) - \mathbf{I} \Vert^2_F }$ & ${\scriptstyle \rvz_k\sim \{\vchi_i(\vmu_\phi, \text{diag}(\bm{\sigma}^2_\phi))\}^{2n}_{i=0} }$ \\
			\midrule
			$\mathcal{L}_{\text{UT-\ac{VAE}*-GP}}$ & ${\scriptstyle \Vert \rvx - \frac{1}{K}\sum^K_{k=1} D_\theta(\rvz_k)\Vert^2_2 + \Vert\vmu_\phi\Vert^2_2+ \Vert \text{diag}(\bm{\sigma}^2_\phi) - \mathbf{I} \Vert^2_F + {\max}(\bm{\sigma}_\phi) \Vert \nabla_{\vmu_\phi} D_\theta(\vmu_\phi)\Vert^2_2 }$ & ${\scriptstyle \rvz_k\sim \{\vchi_i(\vmu_\phi, \text{diag}(\bm{\sigma}^2_\phi))\}^{2n}_{i=0} }$ \\
			\midrule
			$\mathcal{L}_{\text{VAE-}\text{full}\hspace{0.02cm} \mSigma_\phi}$ & ${\scriptstyle \frac{1}{K}\sum^K_{k=1}  \Vert \rvx - D_\theta(\rvz_k)\Vert^2_2 + \Vert\vmu_\phi\Vert^2_2 + \text{tr}(\mSigma_\phi) - 2\text{tr}(\log \bm{L}_\phi)}$ & ${\scriptstyle \rvz_k = \vmu_\phi + \bm{L}_\phi\bm{\epsilon}_k, \hspace{.1cm}   \hspace{.1cm} \bm{\epsilon}_k\sim\mathcal{N}(\mathbf{0}, \mathbf{I})}$ \\
			\midrule
			$\mathcal{L}_{\text{VAE-}\text{full}\hspace{0.02cm} \mSigma_\phi\text{-GP}}$ & ${\scriptstyle \frac{1}{K}\sum^K_{k=1}  \Vert \rvx - D_\theta(\rvz_k)\Vert^2_2 + \Vert\vmu_\phi\Vert^2_2 + \text{tr}(\mSigma_\phi) - 2\text{tr}(\log \bm{L}_\phi) + \lambda_{\max}(\mSigma_\phi) \Vert \nabla_{\vmu_\phi} D_\theta(\vmu_\phi)\Vert^2_2}$ & ${\scriptstyle \rvz_k = \vmu_\phi + \bm{L}_\phi\bm{\epsilon}_k, \hspace{.1cm}   \hspace{.1cm} \bm{\epsilon}_k\sim\mathcal{N}(\mathbf{0}, \mathbf{I})}$ \\
			\midrule
			$\mathcal{L}_{\text{UT-\ac{VAE}-}\text{full}\hspace{0.02cm} \mSigma_\phi}$ & ${\scriptstyle \Vert \rvx - \frac{1}{K}\sum^K_{k=1} D_\theta(\rvz_k)\Vert^2_2 + \Vert\vmu_\phi\Vert^2_2  + \text{tr}(\mSigma_\phi) - 2\text{tr}(\log \bm{L}_\phi) }$ & ${\scriptstyle \rvz_k\sim \{\vchi_i(\vmu_\phi, \mSigma_\phi)\}^{2n}_{i=0} }$ \\
			\midrule
			$\mathcal{L}_{\text{UT-\ac{VAE}-}\text{full}\hspace{0.02cm} \mSigma_\phi\text{-GP}}$ & ${\scriptstyle \Vert \rvx - \frac{1}{K}\sum^K_{k=1} D_\theta(\rvz_k)\Vert^2_2 + \Vert\vmu_\phi\Vert^2_2 + \text{tr}(\mSigma_\phi) - 2\text{tr}(\log \bm{L}_\phi) + \lambda_{\max}(\mSigma_\phi) \Vert \nabla_{\vmu_\phi} D_\theta(\vmu_\phi)\Vert^2_2 }$ & ${\scriptstyle \rvz_k\sim \{\vchi_i(\vmu_\phi, \mSigma_\phi)\}^{2n}_{i=0} }$ \\
			\midrule
			$\mathcal{L}_{\text{VAE*-}\text{full}\hspace{0.02cm} \mSigma_\phi}$ & ${\scriptstyle \frac{1}{K}\sum^K_{k=1}  \Vert \rvx - D_\theta(\rvz_k)\Vert^2_2 + \Vert\vmu_\phi\Vert^2_2 + \Vert \bm{L}_\phi - \mathbf{I} \Vert^2_F }$ & ${\scriptstyle \rvz_k = \vmu_\phi + \bm{L}_\phi\bm{\epsilon}_k, \hspace{.1cm}   \hspace{.1cm} \bm{\epsilon}_k\sim\mathcal{N}(\mathbf{0}, \mathbf{I})}$ \\
			\midrule
			$\mathcal{L}_{\text{VAE*-}\text{full}\hspace{0.02cm} \mSigma_\phi\text{-GP}}$ & ${\scriptstyle \frac{1}{K}\sum^K_{k=1}  \Vert \rvx - D_\theta(\rvz_k)\Vert^2_2 + \Vert\vmu_\phi\Vert^2_2 + \Vert \bm{L}_\phi - \mathbf{I} \Vert^2_F + \lambda_{\max}(\mSigma_\phi) \Vert \nabla_{\vmu_\phi} D_\theta(\vmu_\phi)\Vert^2_2}$ & ${\scriptstyle \rvz_k = \vmu_\phi + \bm{L}_\phi\bm{\epsilon}_k, \hspace{.1cm}   \hspace{.1cm} \bm{\epsilon}_k\sim\mathcal{N}(\mathbf{0}, \mathbf{I})}$ \\
			\midrule
			$\mathcal{L}_{\text{UT-VAE*-}\text{full}\hspace{0.02cm} \mSigma_\phi}$ & ${\scriptstyle \Vert \rvx - \frac{1}{K}\sum^K_{k=1} D_\theta(\rvz_k)\Vert^2_2 + \Vert\vmu_\phi\Vert^2_2 + \Vert \bm{L}_\phi - \mathbf{I} \Vert^2_F }$ & ${\scriptstyle \rvz_k\sim \{\vchi_i(\vmu_\phi, \mSigma_\phi)\}^{2n}_{i=0}}$ \\
			\midrule
			$\mathcal{L}_{\text{UT-VAE*-}\text{full}\hspace{0.02cm} \mSigma_\phi\text{-GP}}$ & ${\scriptstyle \Vert \rvx - \frac{1}{K}\sum^K_{k=1} D_\theta(\rvz_k)\Vert^2_2 + \Vert\vmu_\phi\Vert^2_2 + \Vert \bm{L}_\phi - \mathbf{I} \Vert^2_F + \lambda_{\max}(\mSigma_\phi) \Vert \nabla_{\vmu_\phi} D_\theta(\vmu_\phi)\Vert^2_2 }$ & ${\scriptstyle \rvz_k\sim \{\vchi_i(\vmu_\phi, \mSigma_\phi)\}^{2n}_{i=0} }$  \\	
			\bottomrule
	\end{tabularx}}
	\label{tab:all_loss_functions}
\end{table}
\begin{table}[h!]
	\centering
	\caption{Full ablation study of the models between the \ac{VAE} and \ac{UAE} (in the UT-\ac{VAE}*-GP row), using the Wasserstein metric denoted by *, unscented transform (UT), and the decoder gradient penalty (GP) components. See the upper half Tab.~\ref{tab:all_loss_functions} for the loss function definitions.}
	\begin{tabularx}{1\textwidth}{l *{10}{Y}}
		\toprule
		& \multicolumn{3}{c}{Fashion-MNIST}  
		& \multicolumn{3}{c}{CIFAR10}  
		& \multicolumn{3}{c}{CelebA}\\
		\cmidrule(lr){2-4} \cmidrule(lr){5-7} \cmidrule(lr){8-10} 
		& Rec. & Sample & Interp. & Rec. & Sample & Interp. & Rec. & Sample & Interp.\\
		\midrule
		\ac{VAE}$_{2\text{x}}$ & 43.66 & 49.01 & 61.03 & 112.7 & 123.2 & 120.6 & 67.29 & 69.92 & 70.00  \\
		\ac{VAE}-GP$_{2\text{x}}$ & 44.17 &	48.63 & 59.58 & 108.9 & 120.3 & 117.5 & 66.94 & 70.16 & 69.77 \\
		UT-\ac{VAE}$_{2\text{x}}$ & 36.25 &	40.30 &	53.10  & 95.70 & 115.4 & 107.4 & 51.61 & 57.42 & 56.56 \\
		UT-\ac{VAE}-GP$_{2\text{x}}$ & 47.77 & 65.24 & 72.43 & 102.6 & 118.6 & 113.1 & 100.4 &	102.2 & 100.3 \\
		\midrule
		\ac{VAE}*$_{2\text{x}}$ & 30.07 & 37.92 & 52.15 & 84.57 & 102.2 & 95.61 & 45.18 & 50.97 & 53.73  \\
		\ac{VAE}*-GP$_{2\text{x}}$ & 29.40 & 38.53 & 53.88 & 85.19 & 103.7 &	96.66 & 41.69 &	48.77 &	51.29 \\
		UT-\ac{VAE}*$_{2\text{x}}$ & \textbf{28.26} & \textbf{36.36} & \textbf{50.69} & 82.17 & 100.7 & 93.80 & 44.32 & 50.33 & 52.40\\
		UT-\ac{VAE}* -GP$_{2\text{x}}$ & 29.29 & 37.59 & 53.69 & \textbf{77.71} & \textbf{96.37} & \textbf{89.71} & \textbf{40.07} & \textbf{47.28} & \textbf{50.51} \\
		\bottomrule
	\end{tabularx}
	\label{tab:ablation}
	\vspace{70pt}
\end{table}

\newpage
\section{ELBO Constraint Derivation}\label{sec:elbo_constraint}
In this section, we complete the derivation of the constraint in Eq.~(\ref{eq:constraint}) to the reformulated version in Eq.~(\ref{eq:predev}). The constraint in Eq.~(\ref{eq:constraint}) can be bounded by the maximum of the decoder output in a single dimension $i$, multiplied by the number of dimensions
\begin{equation}
	\Vert D_\theta(\rvz_1) - D_\theta(\rvz_2) \Vert_p \leq \dim(\rvx) \cdot \sup_i \{\Vert d_i(\rvz_1) - d_i(\rvz_2) \Vert_p \} < \epsilon \ . 
\end{equation}
Using the mean value theorem, the term $\sup_i \{\Vert d_i(\rvz_1) - d_i(\rvz_2) \Vert_p \}$ can be reduced to
\begin{align}
	\sup_i \{\Vert \nabla_t d_i((1-t)\rvz_1+t \rvz_2)\Vert_p \cdot \Vert \rvz_1 - \rvz_2 \Vert_p \} &< \epsilon \ ,
\end{align}
Since $\rvz_1$ and $\rvz_2$ are arbitrary, the first part can be simplified and generalized over all dimensions while separating the overall product using the Cauchy-Schwarz inequality
\begin{align}
	\sup_i \{\Vert \nabla_\rvz d_i(\rvz)\Vert_p \cdot \Vert \rvz_1 - \rvz_2 \Vert_p \} &< \epsilon \ \\
	\sup \{\Vert \nabla_\rvz D_\theta(\rvz)\Vert_p \} \cdot \sup \{\Vert \rvz_1 - \rvz_2 \Vert_p \} &< \epsilon \ ,
\end{align}
obtaining the form in Eq.~(\ref{eq:predev}).

\section{Full-Covariance Posterior}\label{sec:ablation_full_cov}
In this section, we aim to investigate the performance of full-covariance posterior models. The non-diagonal posterior representation is naturally supported by the unscented transform and common in filtering. However, it is seldom in \ac{VAE}s -- one of the key ingredients of the standard \ac{VAE} model is its diagonal Gaussian posterior approximation. The induced orthogonality can implicitly have positive effects on the structure of the latent space and the decoder~\citep{zietlow2021demystifying,rolinek2019variational}, but such effects highly depend on implicit biases present in the dataset~\citep{zietlow2021demystifying}. Furthermore, the diagonal posterior together with the \ac{KL} regularization allows for pruning unnecessary latent dimensions, also known as desired posterior collapse~\citep{dai2020usual}. A full-covariance posterior does not have such implicit biases and pruning properties, but it can have a positive effect on the optimization of the variational objective, as it connects otherwise disconnected global optima~\citep{dai2018connections}. Furthermore, it allows for modeling correlations in the posterior. We are not aware of a work successfully employing a full-covariance posterior.

The full-covariance representation can be practically realized by predicting $n$-dimensional standard deviations $\bm{\sigma}_\phi$ as well as $n(n-1)/2$-dimensional correlation factors $\bm{r}_\phi$ (followed by a $\tanh$ projection into the valid $[-1, 1]$ range), and building the lower triangular covariance matrix\footnote{In the $3$-dimensional case: $\bm{L}_\phi=[\sigma_1\quad0\quad0;\quad r_1\sigma_2\sigma_1\quad\sigma_2\quad0;\quad r_2\sigma_3\sigma_1\quad r_3\sigma_3\sigma_2\quad\sigma_3]$.}~$\bm{L}_\phi$. In this way, the full-covariance matrix $\mSigma_\phi=\bm{L}_\phi\bm{L}_\phi^T$ is ensured to be symmetric and positive semi-definite.
 
The results of the full-covariance models are shown in the bottom half of Tab.~\ref{tab:full_covariance_ablation}. In all KL divergence instances, the performance of the models regresses significantly compared to their counterparts in Tab.~\ref{tab:ablation}. This indicates that, despite its theoretical potential to connect disconnected global optima of the optimization objective, a non-diagonal latent space is nevertheless difficult to train with KL divergence, regardless of the sampling method. However, the Wasserstein metric models receive a surprising performance boost. In some cases, they significantly outperform the models from Tab.~\ref{tab:ablation} on Fashion-MNIST and CelebA while achieving similar results on CIFAR10, which has less structure in its input data. It is evident that the Wasserstein metric and potentially its lower posterior variance can enable a successful utilization of correlations in the posterior. 

\begin{table}[]
	\centering
	\caption{Ablation study of the models in Tab.~\ref{tab:ablation} in a full-covariance setting. See Tab.~\ref{tab:all_loss_functions} for the loss function definitions.}
		\begin{tabularx}{1\textwidth}{l *{10}{Y}}
			\toprule
			& \multicolumn{3}{c}{Fashion-MNIST}  
			& \multicolumn{3}{c}{CIFAR10}  
			& \multicolumn{3}{c}{CelebA}\\
			\cmidrule(lr){2-4} \cmidrule(lr){5-7} \cmidrule(lr){8-10} 
			& Rec. & Sample & Interp. & Rec. & Sample & Interp. & Rec. & Sample & Interp.\\
			\midrule
			\ac{VAE}-$\text{full}\hspace{0.05cm} \mSigma_\phi$$_{2\text{x}}$& 79.01 & 83.15 & 91.01 & 123.8 &	132.6 & 130.2 & 99.72 &	100.9 &	99.96 \\
			\ac{VAE}-$\text{full}\hspace{0.05cm} \mSigma_\phi$-GP$_{2\text{x}}$ & 180.0 & 181.5 & 184.4 & 158.3 & 165.8 & 164.0 & 244.2 & 244.6 & 241.8 \\
			UT-\ac{VAE}-$\text{full}\hspace{0.05cm} \mSigma_\phi$$_{2\text{x}}$ & 57.93 & 58.87 & 64.86 &129.6 & 141.2 & 138.2 & 132.1 & 132.4 & 136.0 \\
			UT-\ac{VAE}-$\text{full}\hspace{0.05cm} \mSigma_\phi$-GP$_{2\text{x}}$ & 133.6 & 136.7 & 136.9 & 208.9 & 217.7 & 212.2 & 303.5 & 304.5 & 303.3 \\
			\midrule
			\ac{VAE}*-$\text{full}\hspace{0.05cm} \mSigma_\phi$$_{2\text{x}}$& 31.16 & 40.99 & 54.73 &85.47 & 103.9 & 96.55 & 42.07 & 48.59 & 50.72 \\
			\ac{VAE}*-$\text{full}\hspace{0.05cm} \mSigma_\phi$-GP$_{2\text{x}}$ & \textbf{19.86} & \textbf{32.71} & 48.84 & 84.19 & 102.9 & 95.63 & 39.69 &	46.76 &	49.70 \\
			UT-\ac{VAE}*-$\text{full}\hspace{0.05cm} \mSigma_\phi$$_{2\text{x}}$ & 21.96 & 34.17 & \textbf{48.32} & \textbf{79.51} & \textbf{98.32} & \textbf{91.82} & 41.54 & 48.32 & 50.29 \\
			UT-\ac{VAE}*-$\text{full}\hspace{0.05cm} \mSigma_\phi$-GP$_{2\text{x}}$ & 24.37 & 34.43 & 51.58 & 82.15 & 100.9 & 94.65 & \textbf{39.48} & \textbf{46.60} & \textbf{48.97} \\
			\bottomrule
	\end{tabularx}
	\label{tab:full_covariance_ablation}
\end{table}
\newpage
\section{Connection to Wasserstein Autoencoders}\label{sec:connection_wae}
Wasserstein-distance autoencoders~\citep{patrini2020sinkhorn,tolstikhin2018wae} use the Wasserstein distance $W_p(q_{\text{agg}}(\rvz), p(\rvz))$ to regularize the aggregated posterior $q_{\text{agg}}(\rvz)$ toward the prior $p(\rvz)=\mathcal{N}(\mathbf{0},\bm{I})$. Instead, we use the Wasserstein distance as a simple regularization of the per-sample posterior. However, there is a simple connection of our posterior regularization to the aggregated posterior regularization. Assuming standard normal posteriors, the aggregated posterior can be represented as a mixture

\begin{equation}
	q_{\text{agg}}(\rvz) = \frac{1}{N} \sum_{n} q(\rvz | \rvx_n) = \frac{1}{N} \sum_{n} \mathcal{N}(\vmu_n, \mSigma_n). 
\end{equation}
In the one-dimensional case (generalizable to multiple dimensions) the mean and variance of the mixture are
\begin{equation}
	\mathcal{N}(\mu_n, \sigma^2_n) \stackrel{i.d.}{=} \mathcal{N}\left(\frac{1}{N}\sum_n \mu_n, \frac{1}{N}\sum_n \left(\sigma^2_n + \mu^2_n\right) - \left(\frac{1}{N}\sum_n \mu_n\right)^2\right)\ .
\end{equation}

Thus, the aggregated posterior Wasserstein metric can be represented as
\begin{equation} \begin{aligned}
		W_2 (q_{\text{agg}}(\rvz), p(\rvz))  &= \left(\frac{1}{N}\sum_n \mu_n\right)^2 + \frac{1}{N}\sum_n \left(\sigma^2_n + \mu^2_n\right) - \left(\frac{1}{N}\sum_n \mu_n\right)^2 - 2\sqrt{\frac{1}{N}\sum_n \left(\sigma^2_n + \mu^2_n\right) - \left(\frac{1}{N}\sum_n \mu_n\right)^2}= \\
		&= \frac{1}{N}\sum_n \left(\sigma^2_n + \mu^2_n\right) - 2\sqrt{\frac{1}{N}\sum_n \left(\sigma^2_n + \mu^2_n\right) - \left(\frac{1}{N}\sum_n \mu_n\right)^2}\ ,
	\end{aligned}
\end{equation}
in the case $p=2$ and while discarding constants. Similarly, the average per-sample posterior metric is
\begin{equation}
	\frac{1}{N}\sum_n  W_2 (q_{\text{pp}}(\rvz|\rvx), p(\rvz))  = \frac{1}{N}\sum_n \left( \mu^2_n + \sigma^2_n - 2\sigma_n \right) = \frac{1}{N}\sum_n \mu^2_n + \frac{1}{N}\sum_n \sigma^2_n - 2\frac{1}{N}\sum_n \sigma_n\ .
\end{equation}

Comparing the aggregated posterior metric with the average per-sample posterior metric yields
\begin{align}
	\frac{1}{N}\sum_n \left(\sigma^2_n + \mu^2_n\right) - 2\sqrt{\frac{1}{N}\sum_n \left(\sigma^2_n + \mu^2_n\right) - \left(\frac{1}{N}\sum_n \mu_n\right)^2} &\leq \frac{1}{N}\sum_n \mu^2_n + \frac{1}{N}\sum_n \sigma^2_n - 2\frac{1}{N}\sum_n \sigma_n \\
	- 2\sqrt{\frac{1}{N}\sum_n \left(\sigma^2_n + \mu^2_n\right) - \left(\frac{1}{N}\sum_n \mu_n\right)^2} &\leq - 2\frac{1}{N}\sum_n \sigma_n \\
	\sqrt{\frac{1}{N}\sum_n \left(\sigma^2_n + \mu^2_n\right) - \left(\frac{1}{N}\sum_n \mu_n\right)^2} &\geq \frac{1}{N}\sum_n \sigma_n \\
	\frac{1}{N}\sum_n \left(\sigma^2_n + \mu^2_n\right) - \left(\frac{1}{N}\sum_n \mu_n\right)^2 &\geq \left(\frac{1}{N}\sum_n \sigma_n\right)^2 \\
	\frac{1}{N}\sum_n \left(\sigma^2_n + \mu^2_n\right) &\geq \left(\frac{1}{N}\sum_n \mu_n\right)^2 + \left(\frac{1}{N}\sum_n \sigma_n\right)^2\ . \label{eq:jensens}
\end{align}
Eq.~(\ref{eq:jensens}) can be regarded as two Jensen's inequalities $f(\E[x])\leq\E[f(x)]$, where $f(x)=x^2$, and $\E[x]=\frac{1}{N}\sum_n x_n$. Thus, the initial inequality holds. It shows that the per-sample posterior Wasserstein metric is an upper bound to the aggregated posterior Wasserstein metric, commonly used in the \ac{WAE}~\cite{tolstikhin2018wae}. Therefore, we can guarantee that the Wasserstein distance of the aggregated posterior to the assumed standard normal prior will not be larger than than the average distance of per-sample posteriors.

In addition to the theoretical argument, in Tab.~\ref{tab:wasserstein_ablation} we offer an empirical comparison of the VAE* with the \ac{WAE}-MMD model from~\cite{tolstikhin2018wae} with aggregated posterior weight $\lambda=10$. We observed that the per-sample posterior regularization significantly outperforms the WAE on Fashion-MNIST and CelebA, while being on par on CIFAR10.

\begin{table}
	\centering
	\caption{Comparison of the Wasserstein autoencoder that utilizes the aggregated posterior Wasserstein metric, and the VAE*, utilizing the per-sample posterior Wasserstein metric in the loss. }
	\begin{tabularx}{1\textwidth}{l *{10}{Y}}
		\toprule
		& \multicolumn{3}{c}{Fashion-MNIST}  
		& \multicolumn{3}{c}{CIFAR10}  
		& \multicolumn{3}{c}{CelebA}\\
		\cmidrule(lr){2-4} \cmidrule(lr){5-7} \cmidrule(lr){8-10} 
		& Rec. & Sample & Interp. & Rec. & Sample & Interp. & Rec. & Sample & Interp.\\
		\midrule
		\ac{WAE}-MMD & 47.58 & 62.44 & 73.94 & 88.31 & \textbf{100.35} &	94.78 & 67.54 & 75.92 & 73.21 \\	
		\ac{VAE}*$_{1\text{x}}$ & \textbf{31.62} & \textbf{38.44} & \textbf{52.33} & \textbf{83.49} & 101.5 & \textbf{94.56} & \textbf{44.69} & \textbf{50.55} & \textbf{53.18} \\			
		\bottomrule
	\end{tabularx}
	\label{tab:wasserstein_ablation}
\end{table}
\newpage
\section{Wasserstein Metric Aggregated Posterior Visualization}\label{sec:agg_post}
\begin{figure}[h!]
	\centering
	\begin{subfigure}{.9\linewidth}
		\centering
		\scalebox{1}{\input{images/agg_posterior_vae.tex}}
		\label{fig:post_dist_vae}
	\end{subfigure}%
	
	\begin{subfigure}{.9\textwidth}
		\centering
		\scalebox{1}{\input{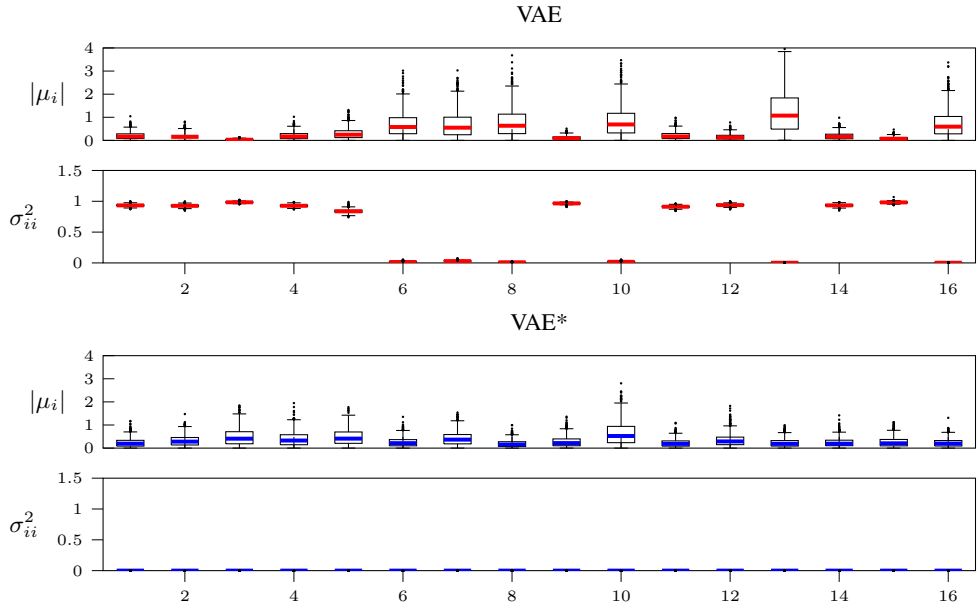}}
		\label{fig:post_dist_vaestar}
	\end{subfigure}
	\caption{Comparison of the distribution of absolute means and variances of 1000 posterior samples for the \ac{VAE}$_{1\text{x}}$ and the \ac{VAE}*$_{1\text{x}}$ models trained with 100 epochs on the CIFAR10 dataset. Top rows show the absolute means and the lower rows the variances of the first 16 dimensions. For the \ac{VAE}*$_{1\text{x}}$ all means differ from zero while the variances are close to zero, whereas for the \ac{VAE}$_{1\text{x}}$, 10 of 16 dimensions are effectively deactivated.}
	\label{fig:agg_posterior}
\end{figure}

In Fig.~\ref{fig:agg_posterior} we present detailed plots on the posterior distributions of \ac{VAE} and \ac{VAE}* for the first 16 dimensions. The \ac{VAE} clearly shows signs of posterior collapse (so-called \textit{polarized regime}~\cite{rolinek2019variational}); we have observed that more than half of the 128 dimensions are nearly equal to the prior. This considerably hurts the generative power of the \ac{VAE} model. In contrast, the \ac{VAE}* model has very low variance in all dimensions, which reflects a nearly deterministic encoder at the end of the training.

\newpage
\section{Qualitative Results on Fashion-MNIST and CIFAR10}\label{sec:qualitative}
\begin{figure*}[h!]
	\centering
	\resizebox{0.9\columnwidth}{!}{%
		\begin{tabularx}{\textwidth}{tmmm}
			& \small Reconstruction & \small Sampling & \small Interpolation \\
			\toprule
			GT & \includegraphics[width=0.35\linewidth]{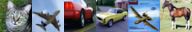} & & \\
			\ac{VAE} & \includegraphics[width=0.35\linewidth]{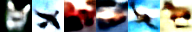} & \includegraphics[width=0.35\linewidth]{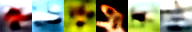} & \includegraphics[width=0.35\linewidth]{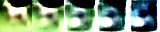}\\
			\ac{RAE} & \includegraphics[width=0.35\linewidth]{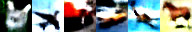} & \includegraphics[width=0.35\linewidth]{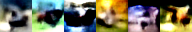} & \includegraphics[width=0.35\linewidth]{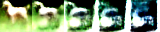} \\
			\ac{UAE} & \includegraphics[width=0.35\linewidth]{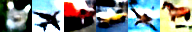} & \includegraphics[width=0.35\linewidth]{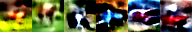} & \includegraphics[width=0.35\linewidth]{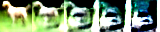} \\
			\bottomrule
	\end{tabularx}}
	\caption{Qualitative results on the CIFAR10 dataset.}
	\label{fig:qualitative_results_cifar10}
\end{figure*}

\begin{figure*}[h!]
	\centering
	\resizebox{0.9\columnwidth}{!}{%
		\begin{tabularx}{\textwidth}{tmmm}
			& \small Reconstruction & \small Sampling & \small Interpolation \\
			\toprule
			GT & \includegraphics[width=0.35\linewidth]{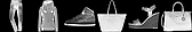} & & \\
			\ac{VAE} & \includegraphics[width=0.35\linewidth]{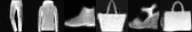} & \includegraphics[width=0.35\linewidth]{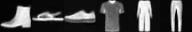} & \includegraphics[width=0.35\linewidth]{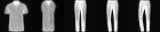}\\
			\ac{RAE} & \includegraphics[width=0.35\linewidth]{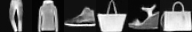} & \includegraphics[width=0.35\linewidth]{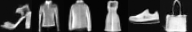} & \includegraphics[width=0.35\linewidth]{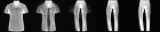} \\
			\ac{UAE} & \includegraphics[width=0.35\linewidth]{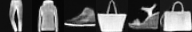} & \includegraphics[width=0.35\linewidth]{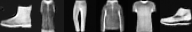} & \includegraphics[width=0.35\linewidth]{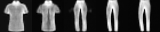} \\
			\bottomrule
	\end{tabularx}}
	\caption{Qualitative results on the Fashion-MNIST dataset.}
	\label{fig:qualitative_results_fashion_mnist}
\end{figure*}

Qualitative results on Fashion-MNIST and CIFAR10 are provided in Fig.~\ref{fig:qualitative_results_cifar10} and Fig.~\ref{fig:qualitative_results_fashion_mnist}. The same setup as in Fig.~\ref{fig:qualitative_results_celeba} is employed. It can be seen that the CIFAR10 images appear considerably richer and sharper, consistent with the results in Tab.~\ref{tab:main_results} and Tab.~\ref{tab:multi-sample}. 

\renewcommand\arraystretch{0.2}
\renewcommand\arraystretch{1}

\renewcommand\arraystretch{0.2}
\renewcommand\arraystretch{1}

%% file: images/agg_posterior_vae.tex
\begin{tikzpicture}

\definecolor{darkgray176}{RGB}{176,176,176}

\begin{groupplot}[group style={group size=1 by 2, vertical sep=0.4cm}]

\nextgroupplot[
title=VAE,
title style={font=\small},
width=375,
height=80,
tick pos=left,
x grid style={darkgray176},
xmin=0.5, xmax=16.5,
xtick style={color=black},
xticklabels={},
y grid style={darkgray176},
ylabel={$|\mu_i|$},
ylabel style={rotate=-90, font=\small},
tick label style={font=\tiny},
ymin=0, ymax=4,
ytick style={color=black}
]
\addplot [black]
table {%
0.75 0.0806790366768837
1.25 0.0806790366768837
1.25 0.283792547881603
0.75 0.283792547881603
0.75 0.0806790366768837
};
\addplot [black]
table {%
1 0.0806790366768837
1 0.000955872237682343
};
\addplot [black]
table {%
1 0.283792547881603
1 0.574116826057434
};
\addplot [black]
table {%
0.875 0.000955872237682343
1.125 0.000955872237682343
};
\addplot [black]
table {%
0.875 0.574116826057434
1.125 0.574116826057434
};
\addplot [black, mark=*, mark size=0.25, mark options={solid,fill opacity=0}, only marks]
table {%
1 0.689721882343292
1 0.612392365932465
1 0.637315034866333
1 0.759713768959045
1 0.817211151123047
1 0.691545486450195
1 0.690334618091583
1 0.597199559211731
1 0.744641900062561
1 0.589905977249146
1 0.775127768516541
1 1.04515480995178
1 0.726924419403076
1 0.624368131160736
1 0.591561198234558
1 0.646612644195557
};
\addplot [black, fill=red, opacity=0.1]
table {%
1.75 0.0736064314842224
2.25 0.0736064314842224
2.25 0.251603007316589
1.75 0.251603007316589
1.75 0.0736064314842224
};
\addplot [black]
table {%
2 0.0736064314842224
2 0.000237822532653809
};
\addplot [black]
table {%
2 0.251603007316589
2 0.514002084732056
};
\addplot [black]
table {%
1.875 0.000237822532653809
2.125 0.000237822532653809
};
\addplot [black]
table {%
1.875 0.514002084732056
2.125 0.514002084732056
};
\addplot [black, mark=*, mark size=0.25, mark options={solid,fill opacity=0}, only marks]
table {%
2 0.573449730873108
2 0.536790132522583
2 0.539019823074341
2 0.794438362121582
2 0.539844989776611
2 0.812301278114319
2 0.606755614280701
2 0.534579277038574
2 0.564966678619385
2 0.52733314037323
2 0.774877309799194
2 0.624336838722229
2 0.534852147102356
2 0.644424676895142
2 0.63555920124054
2 0.626432061195374
2 0.557939410209656
2 0.521744787693024
2 0.546449780464172
2 0.578014969825745
2 0.680374264717102
2 0.520491063594818
2 0.625580728054047
2 0.555866479873657
2 0.663727164268494
2 0.523042321205139
2 0.521876752376556
2 0.581902027130127
};
\addplot [black]
table {%
2.75 0.0136536099016666
3.25 0.0136536099016666
3.25 0.0481022968888283
2.75 0.0481022968888283
2.75 0.0136536099016666
};
\addplot [black]
table {%
3 0.0136536099016666
3 7.01099634170532e-06
};
\addplot [black]
table {%
3 0.0481022968888283
3 0.0993736907839775
};
\addplot [black]
table {%
2.875 7.01099634170532e-06
3.125 7.01099634170532e-06
};
\addplot [black]
table {%
2.875 0.0993736907839775
3.125 0.0993736907839775
};
\addplot [black, mark=*, mark size=0.25, mark options={solid,fill opacity=0}, only marks]
table {%
3 0.128965735435486
3 0.119889073073864
3 0.123912535607815
3 0.117545627057552
3 0.125446170568466
3 0.10677146166563
3 0.115814574062824
3 0.110954143106937
3 0.120562516152859
3 0.111505500972271
3 0.109271861612797
3 0.10384913533926
3 0.108885131776333
3 0.111016504466534
3 0.101758860051632
3 0.136893600225449
3 0.122618414461613
3 0.114220894873142
3 0.100228913128376
3 0.126494646072388
};
\addplot [black]
table {%
3.75 0.0773382224142551
4.25 0.0773382224142551
4.25 0.294164851307869
3.75 0.294164851307869
3.75 0.0773382224142551
};
\addplot [black]
table {%
4 0.0773382224142551
4 0.000203793868422508
};
\addplot [black]
table {%
4 0.294164851307869
4 0.61155366897583
};
\addplot [black]
table {%
3.875 0.000203793868422508
4.125 0.000203793868422508
};
\addplot [black]
table {%
3.875 0.61155366897583
4.125 0.61155366897583
};
\addplot [black, mark=*, mark size=0.25, mark options={solid,fill opacity=0}, only marks]
table {%
4 0.681562066078186
4 0.676918148994446
4 0.716007351875305
4 0.750130474567413
4 0.835926234722137
4 0.732835710048676
4 0.686358213424683
4 0.755757987499237
4 1.02161359786987
4 0.633471250534058
4 0.672137200832367
4 0.639285326004028
};
\addplot [black]
table {%
4.75 0.120872803032398
5.25 0.120872803032398
5.25 0.417646050453186
4.75 0.417646050453186
4.75 0.120872803032398
};
\addplot [black]
table {%
5 0.120872803032398
5 0.000293374061584473
};
\addplot [black]
table {%
5 0.417646050453186
5 0.86076283454895
};
\addplot [black]
table {%
4.875 0.000293374061584473
5.125 0.000293374061584473
};
\addplot [black]
table {%
4.875 0.86076283454895
5.125 0.86076283454895
};
\addplot [black, mark=*, mark size=0.25, mark options={solid,fill opacity=0}, only marks]
table {%
5 0.921664655208588
5 1.31225633621216
5 1.02117323875427
5 1.0135098695755
5 0.871931850910187
5 0.881972312927246
5 1.09942531585693
5 1.27868044376373
5 0.872107446193695
5 0.961603343486786
5 1.11731219291687
5 0.911827802658081
5 0.932493686676025
5 0.866916656494141
5 0.997160911560059
5 1.10888409614563
5 1.21651983261108
5 0.908014714717865
5 0.971095561981201
5 1.26390588283539
5 1.22739410400391
5 0.864666104316711
5 0.914551377296448
};
\addplot [black]
table {%
5.75 0.289855383336544
6.25 0.289855383336544
6.25 0.983043760061264
5.75 0.983043760061264
5.75 0.289855383336544
};
\addplot [black]
table {%
6 0.289855383336544
6 0.000241219997406006
};
\addplot [black]
table {%
6 0.983043760061264
6 2.0102105140686
};
\addplot [black]
table {%
5.875 0.000241219997406006
6.125 0.000241219997406006
};
\addplot [black]
table {%
5.875 2.0102105140686
6.125 2.0102105140686
};
\addplot [black, mark=*, mark size=0.25, mark options={solid,fill opacity=0}, only marks]
table {%
6 2.76762938499451
6 2.1081600189209
6 2.17241716384888
6 2.12833619117737
6 2.04552721977234
6 2.1437885761261
6 2.59159445762634
6 2.20354843139648
6 3.01815915107727
6 2.5159125328064
6 2.38737034797668
6 2.91047716140747
6 2.36151695251465
6 2.20034575462341
6 2.12811350822449
6 2.67875933647156
6 2.48168158531189
};
\addplot [black]
table {%
6.75 0.245258077979088
7.25 0.245258077979088
7.25 1.00414845347404
6.75 1.00414845347404
6.75 0.245258077979088
};
\addplot [black]
table {%
7 0.245258077979088
7 0.000525414943695068
};
\addplot [black]
table {%
7 1.00414845347404
7 2.13029670715332
};
\addplot [black]
table {%
6.875 0.000525414943695068
7.125 0.000525414943695068
};
\addplot [black]
table {%
6.875 2.13029670715332
7.125 2.13029670715332
};
\addplot [black, mark=*, mark size=0.25, mark options={solid,fill opacity=0}, only marks]
table {%
7 2.4780387878418
7 2.34197473526001
7 2.21096897125244
7 2.17722463607788
7 2.26360273361206
7 2.16842222213745
7 2.18871593475342
7 2.67783832550049
7 2.40699577331543
7 2.49911546707153
7 2.42178726196289
7 2.46299695968628
7 2.70552778244019
7 2.55049085617065
7 3.02794933319092
7 2.45162153244019
7 2.41910076141357
7 2.22737693786621
7 2.47228145599365
7 2.60113191604614
7 2.50130844116211
7 2.3466944694519
7 2.14713668823242
7 4.39653778076172
7 2.45330667495728
};
\addplot [black]
table {%
7.75 0.293252855539322
8.25 0.293252855539322
8.25 1.13534480333328
7.75 1.13534480333328
7.75 0.293252855539322
};
\addplot [black]
table {%
8 0.293252855539322
8 0.00140023231506348
};
\addplot [black]
table {%
8 1.13534480333328
8 2.35253882408142
};
\addplot [black]
table {%
7.875 0.00140023231506348
8.125 0.00140023231506348
};
\addplot [black]
table {%
7.875 2.35253882408142
8.125 2.35253882408142
};
\addplot [black, mark=*, mark size=0.25, mark options={solid,fill opacity=0}, only marks]
table {%
8 2.87831664085388
8 2.52824711799622
8 2.46194386482239
8 2.51628160476685
8 2.55802321434021
8 2.56866192817688
8 2.66454458236694
8 3.68116545677185
8 2.7021312713623
8 2.47949028015137
8 2.93174290657043
8 2.43613767623901
8 2.43273782730103
8 2.73081851005554
8 3.37845492362976
8 2.67719388008118
8 3.11033654212952
8 2.62577795982361
};
\addplot [black]
table {%
8.75 0.0425980091094971
9.25 0.0425980091094971
9.25 0.154623780399561
8.75 0.154623780399561
8.75 0.0425980091094971
};
\addplot [black]
table {%
9 0.0425980091094971
9 0.000490069389343262
};
\addplot [black]
table {%
9 0.154623780399561
9 0.320447981357574
};
\addplot [black]
table {%
8.875 0.000490069389343262
9.125 0.000490069389343262
};
\addplot [black]
table {%
8.875 0.320447981357574
9.125 0.320447981357574
};
\addplot [black, mark=*, mark size=0.25, mark options={solid,fill opacity=0}, only marks]
table {%
9 0.325877726078033
9 0.345993906259537
9 0.339199125766754
9 0.34694892168045
9 0.333238244056702
9 0.394869446754456
9 0.332684099674225
9 0.379367291927338
9 0.392711788415909
9 0.410733938217163
9 0.356186926364899
9 0.424974769353867
9 0.357775449752808
9 0.420538544654846
9 0.510139763355255
9 0.415664434432983
9 0.362185448408127
};
\addplot [black]
table {%
9.75 0.323298327624798
10.25 0.323298327624798
10.25 1.17231371998787
9.75 1.17231371998787
9.75 0.323298327624798
};
\addplot [black]
table {%
10 0.323298327624798
10 0.00254517793655396
};
\addplot [black]
table {%
10 1.17231371998787
10 2.44088006019592
};
\addplot [black]
table {%
9.875 0.00254517793655396
10.125 0.00254517793655396
};
\addplot [black]
table {%
9.875 2.44088006019592
10.125 2.44088006019592
};
\addplot [black, mark=*, mark size=0.25, mark options={solid,fill opacity=0}, only marks]
table {%
10 3.47197318077087
10 2.82383728027344
10 2.56359601020813
10 3.0975866317749
10 2.52027416229248
10 2.94634628295898
10 2.77935528755188
10 3.21244764328003
10 3.33468699455261
10 2.56963968276978
10 2.71547150611877
10 2.85427331924438
10 2.61469221115112
10 3.02128839492798
10 2.57225584983826
10 2.53072667121887
10 2.93852186203003
10 2.4921658039093
10 3.01586961746216
10 3.31807518005371
10 2.82638096809387
};
\addplot [black]
table {%
10.75 0.0814711898565292
11.25 0.0814711898565292
11.25 0.298281908035278
10.75 0.298281908035278
10.75 0.0814711898565292
};
\addplot [black]
table {%
11 0.0814711898565292
11 9.1850757598877e-05
};
\addplot [black]
table {%
11 0.298281908035278
11 0.619035303592682
};
\addplot [black]
table {%
10.875 9.1850757598877e-05
11.125 9.1850757598877e-05
};
\addplot [black]
table {%
10.875 0.619035303592682
11.125 0.619035303592682
};
\addplot [black, mark=*, mark size=0.25, mark options={solid,fill opacity=0}, only marks]
table {%
11 0.863303959369659
11 0.663670659065247
11 0.747396409511566
11 0.631666600704193
11 0.726064622402191
11 0.629197180271149
11 0.957290709018707
11 0.775896370410919
11 0.982379376888275
11 0.656072080135345
11 0.63323038816452
11 0.65418404340744
11 0.634563624858856
11 0.729544639587402
11 0.670127928256989
11 0.670799314975739
11 0.65960818529129
11 0.846099436283112
11 0.657344996929169
};
\addplot [black]
table {%
11.75 0.0602009072899818
12.25 0.0602009072899818
12.25 0.220527827739716
11.75 0.220527827739716
11.75 0.0602009072899818
};
\addplot [black]
table {%
12 0.0602009072899818
12 0.000464826822280884
};
\addplot [black]
table {%
12 0.220527827739716
12 0.459512650966644
};
\addplot [black]
table {%
11.875 0.000464826822280884
12.125 0.000464826822280884
};
\addplot [black]
table {%
11.875 0.459512650966644
12.125 0.459512650966644
};
\addplot [black, mark=*, mark size=0.25, mark options={solid,fill opacity=0}, only marks]
table {%
12 0.470537900924683
12 0.513666749000549
12 0.489715665578842
12 0.608981966972351
12 0.578302383422852
12 0.4855597615242
12 0.554671406745911
12 0.469010859727859
12 0.470750480890274
12 0.590322136878967
12 0.618508458137512
12 0.505233764648438
12 0.551806449890137
12 0.776506662368774
12 0.643760502338409
12 0.479035764932632
};
\addplot [black]
table {%
12.75 0.488819807767868
13.25 0.488819807767868
13.25 1.83706468343735
12.75 1.83706468343735
12.75 0.488819807767868
};
\addplot [black]
table {%
13 0.488819807767868
13 0.00854873657226562
};
\addplot [black]
table {%
13 1.83706468343735
13 3.84348058700562
};
\addplot [black]
table {%
12.875 0.00854873657226562
13.125 0.00854873657226562
};
\addplot [black]
table {%
12.875 3.84348058700562
13.125 3.84348058700562
};
\addplot [black, mark=*, mark size=0.25, mark options={solid,fill opacity=0}, only marks]
table {%
13 4.52793788909912
13 4.06777334213257
13 4.34385395050049
13 4.02261590957642
13 4.64348602294922
13 4.52875089645386
13 3.9543981552124
13 4.55837535858154
};
\addplot [black]
table {%
13.75 0.0725082419812679
14.25 0.0725082419812679
14.25 0.268743298947811
13.75 0.268743298947811
13.75 0.0725082419812679
};
\addplot [black]
table {%
14 0.0725082419812679
14 2.14576721191406e-06
};
\addplot [black]
table {%
14 0.268743298947811
14 0.556621074676514
};
\addplot [black]
table {%
13.875 2.14576721191406e-06
14.125 2.14576721191406e-06
};
\addplot [black]
table {%
13.875 0.556621074676514
14.125 0.556621074676514
};
\addplot [black, mark=*, mark size=0.25, mark options={solid,fill opacity=0}, only marks]
table {%
14 0.574469923973083
14 0.692597389221191
14 0.565234303474426
14 0.619888007640839
14 0.728854179382324
14 0.724312305450439
14 0.583072304725647
14 0.617920756340027
14 0.733492016792297
14 0.67861008644104
14 0.603921413421631
14 0.67512035369873
14 0.690704941749573
14 0.981029987335205
14 0.6055588722229
14 0.624516725540161
14 0.66233503818512
14 0.618499994277954
};
\addplot [black]
table {%
14.75 0.0353698879480362
15.25 0.0353698879480362
15.25 0.123730558902025
14.75 0.123730558902025
14.75 0.0353698879480362
};
\addplot [black]
table {%
15 0.0353698879480362
15 8.41021537780762e-05
};
\addplot [black]
table {%
15 0.123730558902025
15 0.253633886575699
};
\addplot [black]
table {%
14.875 8.41021537780762e-05
15.125 8.41021537780762e-05
};
\addplot [black]
table {%
14.875 0.253633886575699
15.125 0.253633886575699
};
\addplot [black, mark=*, mark size=0.25, mark options={solid,fill opacity=0}, only marks]
table {%
15 0.269681215286255
15 0.265708744525909
15 0.269863843917847
15 0.269977152347565
15 0.28687259554863
15 0.285490870475769
15 0.355886936187744
15 0.302529692649841
15 0.264161884784698
15 0.296409845352173
15 0.259596526622772
15 0.257167100906372
15 0.263316810131073
15 0.301360428333282
15 0.307943046092987
15 0.328578650951385
15 0.335761547088623
15 0.272245079278946
15 0.325422644615173
15 0.27416118979454
15 0.469170957803726
15 0.343176752328873
15 0.276095628738403
};
\addplot [black]
table {%
15.75 0.284347176551819
16.25 0.284347176551819
16.25 1.03501403331757
15.75 1.03501403331757
15.75 0.284347176551819
};
\addplot [black]
table {%
16 0.284347176551819
16 9.13143157958984e-05
};
\addplot [black]
table {%
16 1.03501403331757
16 2.1576828956604
};
\addplot [black]
table {%
15.875 9.13143157958984e-05
16.125 9.13143157958984e-05
};
\addplot [black]
table {%
15.875 2.1576828956604
16.125 2.1576828956604
};
\addplot [black, mark=*, mark size=0.25, mark options={solid,fill opacity=0}, only marks]
table {%
16 2.23218297958374
16 2.41422605514526
16 2.45976161956787
16 2.25724840164185
16 2.45709943771362
16 2.68212080001831
16 2.58533668518066
16 2.2269434928894
16 2.25842237472534
16 2.18361854553223
16 2.331130027771
16 2.42426323890686
16 2.45514249801636
16 2.54787683486938
16 2.73232269287109
16 3.20482015609741
16 2.56268715858459
16 2.73362016677856
16 3.37259483337402
16 2.19328451156616
16 2.37388372421265
16 2.33868408203125
16 2.42725586891174
16 2.23115062713623
16 2.1897988319397
16 2.29202389717102
16 3.18933916091919
16 2.56621503829956
};
\addplot [line width=1.5pt, red]
table {%
0.75 0.16552822291851
1.25 0.16552822291851
};
\addplot [line width=1.5pt, red]
table {%
1.75 0.152935981750488
2.25 0.152935981750488
};
\addplot [line width=1.5pt, red]
table {%
2.75 0.0290569923818111
3.25 0.0290569923818111
};
\addplot [line width=1.5pt, red]
table {%
3.75 0.169895604252815
4.25 0.169895604252815
};
\addplot [line width=1.5pt, red]
table {%
4.75 0.2469642162323
5.25 0.2469642162323
};
\addplot [line width=1.5pt, red]
table {%
5.75 0.579681873321533
6.25 0.579681873321533
};
\addplot [line width=1.5pt, red]
table {%
6.75 0.549811661243439
7.25 0.549811661243439
};
\addplot [line width=1.5pt, red]
table {%
7.75 0.633782088756561
8.25 0.633782088756561
};
\addplot [line width=1.5pt, red]
table {%
8.75 0.0946638286113739
9.25 0.0946638286113739
};
\addplot [line width=1.5pt, red]
table {%
9.75 0.689428627490997
10.25 0.689428627490997
};
\addplot [line width=1.5pt, red]
table {%
10.75 0.171936124563217
11.25 0.171936124563217
};
\addplot [line width=1.5pt, red]
table {%
11.75 0.128201961517334
12.25 0.128201961517334
};
\addplot [line width=1.5pt, red]
table {%
12.75 1.07244449853897
13.25 1.07244449853897
};
\addplot [line width=1.5pt, red]
table {%
13.75 0.159545123577118
14.25 0.159545123577118
};
\addplot [line width=1.5pt, red]
table {%
14.75 0.073772519826889
15.25 0.073772519826889
};
\addplot [line width=1.5pt, red]
table {%
15.75 0.594440281391144
16.25 0.594440281391144
};

\nextgroupplot[
width=375,
height=80,
scaled x ticks=manual:{}{\pgfmathparse{#1}},
tick align=outside,
tick pos=left,
x grid style={darkgray176},
xmin=0.5, xmax=16.5,
xtick style={color=black},
y grid style={darkgray176},
ymin=0, ymax=1.5,
ytick style={color=black},
ylabel={$\sigma^2_{ii}$},
ylabel style={rotate=-90, font=\small},
tick label style={font=\tiny},]
\addplot [black]
table {%
0.75 0.922216549515724
1.25 0.922216549515724
1.25 0.944075927138329
0.75 0.944075927138329
0.75 0.922216549515724
};
\addplot [black]
table {%
1 0.922216549515724
1 0.891348302364349
};
\addplot [black]
table {%
1 0.944075927138329
1 0.976391911506653
};
\addplot [black]
table {%
0.875 0.891348302364349
1.125 0.891348302364349
};
\addplot [black]
table {%
0.875 0.976391911506653
1.125 0.976391911506653
};
\addplot [black, mark=*, mark size=0.25, mark options={solid,fill opacity=0}, only marks]
table {%
1 0.887225329875946
1 0.887838661670685
1 0.887996971607208
1 0.884337782859802
1 0.883111417293549
1 0.881131112575531
1 0.887277841567993
1 0.874001741409302
1 0.870821535587311
1 0.996436476707458
1 0.983552277088165
1 0.995362997055054
1 0.995758414268494
1 0.992599844932556
1 0.97800749540329
1 0.986094355583191
1 0.989888250827789
1 0.987679958343506
1 0.978545010089874
1 0.979213833808899
1 0.977189779281616
1 0.985212802886963
1 0.987758219242096
1 0.979888021945953
1 0.99142187833786
};
\addplot [black]
table {%
1.75 0.914612665772438
2.25 0.914612665772438
2.25 0.937331452965736
1.75 0.937331452965736
1.75 0.914612665772438
};
\addplot [black]
table {%
2 0.914612665772438
2 0.880770623683929
};
\addplot [black]
table {%
2 0.937331452965736
2 0.971346497535706
};
\addplot [black]
table {%
1.875 0.880770623683929
2.125 0.880770623683929
};
\addplot [black]
table {%
1.875 0.971346497535706
2.125 0.971346497535706
};
\addplot [black, mark=*, mark size=0.25, mark options={solid,fill opacity=0}, only marks]
table {%
2 0.869182825088501
2 0.873669803142548
2 0.846836388111115
2 0.879265487194061
2 0.877347230911255
2 0.878649652004242
2 0.868524193763733
2 0.873791933059692
2 0.860239446163177
2 0.879296958446503
2 0.879790723323822
2 0.854935526847839
2 0.868503093719482
2 0.879704892635345
2 0.861942291259766
2 0.878432273864746
2 0.867769956588745
2 0.877511143684387
2 0.971757352352142
2 0.984951019287109
2 0.983650684356689
2 0.976280450820923
2 0.979175209999084
2 0.975992858409882
2 0.97281277179718
2 0.980562031269073
2 0.988499760627747
2 0.998892486095428
2 0.97229677438736
2 0.981840431690216
2 0.973187744617462
2 0.971804082393646
2 0.974645853042603
};
\addplot [black]
table {%
2.75 0.97803895175457
3.25 0.97803895175457
3.25 0.989598140120506
2.75 0.989598140120506
2.75 0.97803895175457
};
\addplot [black]
table {%
3 0.97803895175457
3 0.961073100566864
};
\addplot [black]
table {%
3 0.989598140120506
3 1.00601553916931
};
\addplot [black]
table {%
2.875 0.961073100566864
3.125 0.961073100566864
};
\addplot [black]
table {%
2.875 1.00601553916931
3.125 1.00601553916931
};
\addplot [black, mark=*, mark size=0.25, mark options={solid,fill opacity=0}, only marks]
table {%
3 0.958911597728729
3 0.959425330162048
3 0.957168757915497
3 0.953772664070129
3 0.960654199123383
3 0.959459483623505
3 0.949616849422455
3 0.959360361099243
3 0.95861405134201
3 1.01016020774841
3 1.01134967803955
3 1.00791764259338
3 1.00794339179993
3 1.01000022888184
3 1.0110422372818
3 1.00984191894531
3 1.01164281368256
3 1.00917685031891
3 1.00699162483215
3 1.01253235340118
3 1.02421343326569
3 1.00740873813629
3 1.00743520259857
3 1.02051043510437
3 1.0072660446167
3 1.00700259208679
};
\addplot [black]
table {%
3.75 0.914619848132133
4.25 0.914619848132133
4.25 0.938798144459724
3.75 0.938798144459724
3.75 0.914619848132133
};
\addplot [black]
table {%
4 0.914619848132133
4 0.883728623390198
};
\addplot [black]
table {%
4 0.938798144459724
4 0.974997699260712
};
\addplot [black]
table {%
3.875 0.883728623390198
4.125 0.883728623390198
};
\addplot [black]
table {%
3.875 0.974997699260712
4.125 0.974997699260712
};
\addplot [black, mark=*, mark size=0.25, mark options={solid,fill opacity=0}, only marks]
table {%
4 0.872149407863617
4 0.867377519607544
4 0.877561569213867
4 0.877228498458862
4 0.98491507768631
4 0.976809144020081
4 0.988570988178253
4 0.985768258571625
4 0.982627093791962
};
\addplot [black]
table {%
4.75 0.819015130400658
5.25 0.819015130400658
5.25 0.855538129806519
4.75 0.855538129806519
4.75 0.819015130400658
};
\addplot [black]
table {%
5 0.819015130400658
5 0.76662266254425
};
\addplot [black]
table {%
5 0.855538129806519
5 0.909625589847565
};
\addplot [black]
table {%
4.875 0.76662266254425
5.125 0.76662266254425
};
\addplot [black]
table {%
4.875 0.909625589847565
5.125 0.909625589847565
};
\addplot [black, mark=*, mark size=0.25, mark options={solid,fill opacity=0}, only marks]
table {%
5 0.762666940689087
5 0.740379154682159
5 0.758615791797638
5 0.753922462463379
5 0.756679713726044
5 0.930087208747864
5 0.937618970870972
5 0.91161322593689
5 0.928261935710907
5 0.918136060237885
5 0.935720026493073
5 0.914770901203156
5 0.973357319831848
5 0.988990604877472
5 0.910352528095245
5 0.915296912193298
5 0.924640357494354
5 0.911494314670563
5 0.935051620006561
5 0.928407549858093
5 0.964352250099182
5 0.919679522514343
5 0.925204694271088
5 0.921449840068817
5 0.928509593009949
5 0.919518947601318
5 0.912754833698273
5 0.921779811382294
5 0.92703515291214
5 0.921240627765656
5 0.935446262359619
5 0.910829484462738
5 0.912197470664978
5 0.913924872875214
5 0.921925067901611
5 0.930940866470337
5 0.949017763137817
5 0.912905931472778
5 0.912209689617157
5 0.91095244884491
5 0.922011196613312
5 0.923725366592407
5 0.92097669839859
5 0.946022748947144
5 0.911334097385406
5 0.920466721057892
5 0.934239149093628
};
\addplot [black]
table {%
5.75 0.0130577669478953
6.25 0.0130577669478953
6.25 0.0172307258471847
5.75 0.0172307258471847
5.75 0.0130577669478953
};
\addplot [black]
table {%
6 0.0130577669478953
6 0.00819305703043938
};
\addplot [black]
table {%
6 0.0172307258471847
6 0.0234508030116558
};
\addplot [black]
table {%
5.875 0.00819305703043938
6.125 0.00819305703043938
};
\addplot [black]
table {%
5.875 0.0234508030116558
6.125 0.0234508030116558
};
\addplot [black, mark=*, mark size=0.25, mark options={solid,fill opacity=0}, only marks]
table {%
6 0.0302657131105661
6 0.0293445326387882
6 0.0244478601962328
6 0.0272105578333139
6 0.0255096238106489
6 0.0237034074962139
6 0.052241425961256
6 0.0286088883876801
6 0.0239674393087626
6 0.0252501219511032
6 0.0239573922008276
6 0.0246015582233667
6 0.0462711788713932
6 0.0284391175955534
6 0.0334959849715233
6 0.0373388677835464
6 0.0256170015782118
6 0.0357509329915047
6 0.0288595706224442
6 0.0431481711566448
6 0.0245907828211784
6 0.0344968810677528
6 0.0262478739023209
6 0.0243436712771654
6 0.0316547565162182
6 0.0243586897850037
6 0.0248259250074625
6 0.024110971018672
6 0.027511365711689
6 0.0391621291637421
6 0.0295591820031404
6 0.0360425561666489
6 0.0275921821594238
6 0.0241136271506548
6 0.0259499996900558
6 0.0345471948385239
6 0.0263180620968342
6 0.0402073375880718
6 0.0258036889135838
6 0.0241438094526529
6 0.0240200851112604
6 0.0257389731705189
6 0.026348764076829
6 0.0357746630907059
6 0.0380203388631344
6 0.0268026608973742
6 0.0283703599125147
6 0.0244245491921902
6 0.0243341419845819
6 0.0343475490808487
6 0.0262793190777302
6 0.0314618349075317
6 0.0467957742512226
6 0.026102913543582
6 0.0290137100964785
6 0.0296075101941824
6 0.0327913053333759
6 0.0256321337074041
6 0.0530656650662422
6 0.0249934382736683
6 0.0289983637630939
6 0.0266076717525721
6 0.0272530559450388
};
\addplot [black]
table {%
6.75 0.0262089096941054
7.25 0.0262089096941054
7.25 0.0334246149286628
6.75 0.0334246149286628
6.75 0.0262089096941054
};
\addplot [black]
table {%
7 0.0262089096941054
7 0.0156638734042645
};
\addplot [black]
table {%
7 0.0334246149286628
7 0.0441788248717785
};
\addplot [black]
table {%
6.875 0.0156638734042645
7.125 0.0156638734042645
};
\addplot [black]
table {%
6.875 0.0441788248717785
7.125 0.0441788248717785
};
\addplot [black, mark=*, mark size=0.25, mark options={solid,fill opacity=0}, only marks]
table {%
7 0.0464569516479969
7 0.0445989295840263
7 0.0745068117976189
7 0.0450202934443951
7 0.0451908968389034
7 0.062379103153944
7 0.0448515117168427
7 0.0613570362329483
7 0.0640076398849487
7 0.0446253083646297
7 0.0601905882358551
7 0.0477015189826488
7 0.0494437739253044
7 0.052124734967947
7 0.0467110686004162
7 0.0554463043808937
7 0.0586449168622494
7 0.0528178140521049
7 0.0513972118496895
7 0.0462312288582325
7 0.0508809126913548
7 0.0492667146027088
7 0.0570638179779053
7 0.045113168656826
7 0.0531975477933884
7 0.0469573251903057
7 0.0453936755657196
7 0.0542393252253532
7 0.0584494099020958
7 0.0443308055400848
7 0.0448902398347855
7 0.0550714731216431
7 0.0498556904494762
7 0.0456661321222782
7 0.0453149154782295
7 0.0509906224906445
7 0.0650272741913795
7 0.0448059067130089
7 0.0458842739462852
7 0.0466699413955212
7 0.0721594020724297
7 0.0450001060962677
7 0.0463021583855152
};
\addplot [black]
table {%
7.75 0.00921737053431571
8.25 0.00921737053431571
8.25 0.0117675978690386
7.75 0.0117675978690386
7.75 0.00921737053431571
};
\addplot [black]
table {%
8 0.00921737053431571
8 0.00594893423840404
};
\addplot [black]
table {%
8 0.0117675978690386
8 0.0155915021896362
};
\addplot [black]
table {%
7.875 0.00594893423840404
8.125 0.00594893423840404
};
\addplot [black]
table {%
7.875 0.0155915021896362
8.125 0.0155915021896362
};
\addplot [black, mark=*, mark size=0.25, mark options={solid,fill opacity=0}, only marks]
table {%
8 0.015876954421401
8 0.0260350871831179
8 0.0209458619356155
8 0.0162671580910683
8 0.0200514290481806
8 0.0213589519262314
8 0.0209239684045315
8 0.0162129662930965
8 0.0159794818609953
8 0.017540916800499
8 0.0161725748330355
8 0.0186878126114607
8 0.0156603995710611
8 0.0196331515908241
8 0.0186576191335917
8 0.0179925002157688
8 0.0175900235772133
8 0.0185857228934765
8 0.0177131276577711
8 0.0203518364578485
8 0.0164606291800737
8 0.0158987045288086
8 0.0208335518836975
8 0.0205907057970762
8 0.0193749703466892
8 0.0163261312991381
8 0.0196303427219391
8 0.0225874595344067
8 0.0156528297811747
8 0.0188698191195726
8 0.0282247941941023
8 0.015645882114768
8 0.0162582006305456
};
\addplot [black]
table {%
8.75 0.959893330931664
9.25 0.959893330931664
9.25 0.97153976559639
8.75 0.97153976559639
8.75 0.959893330931664
};
\addplot [black]
table {%
9 0.959893330931664
9 0.942771375179291
};
\addplot [black]
table {%
9 0.97153976559639
9 0.987572431564331
};
\addplot [black]
table {%
8.875 0.942771375179291
9.125 0.942771375179291
};
\addplot [black]
table {%
8.875 0.987572431564331
9.125 0.987572431564331
};
\addplot [black, mark=*, mark size=0.25, mark options={solid,fill opacity=0}, only marks]
table {%
9 0.935069143772125
9 0.94215726852417
9 0.941810429096222
9 0.924684762954712
9 0.941769421100616
9 0.914914727210999
9 0.941144049167633
9 0.907658934593201
9 0.939240992069244
9 0.930076718330383
9 0.938607692718506
9 0.93784773349762
9 0.922561168670654
9 0.940834820270538
9 0.932276785373688
9 0.909709930419922
9 0.92450225353241
9 0.929535210132599
9 0.93182909488678
9 0.936250746250153
9 0.992760121822357
9 1.00003409385681
9 0.989547610282898
9 0.98917031288147
9 0.994069337844849
9 0.989781677722931
9 0.991351544857025
9 0.993064105510712
9 0.98934805393219
9 0.99370139837265
9 0.992083966732025
};
\addplot [black]
table {%
9.75 0.0159338489174843
10.25 0.0159338489174843
10.25 0.020520347636193
9.75 0.020520347636193
9.75 0.0159338489174843
};
\addplot [black]
table {%
10 0.0159338489174843
10 0.0102072432637215
};
\addplot [black]
table {%
10 0.020520347636193
10 0.0271859802305698
};
\addplot [black]
table {%
9.875 0.0102072432637215
10.125 0.0102072432637215
};
\addplot [black]
table {%
9.875 0.0271859802305698
10.125 0.0271859802305698
};
\addplot [black, mark=*, mark size=0.25, mark options={solid,fill opacity=0}, only marks]
table {%
10 0.0288666654378176
10 0.0332834608852863
10 0.0275136791169643
10 0.0527296997606754
10 0.027673177421093
10 0.0438388884067535
10 0.0317303538322449
10 0.0433799363672733
10 0.048632800579071
10 0.035970363765955
10 0.0295479353517294
10 0.0360890105366707
10 0.0304795075207949
10 0.0362132452428341
10 0.0299880430102348
10 0.0433423854410648
10 0.0298315305262804
10 0.0327630415558815
10 0.0360120683908463
10 0.0415176041424274
10 0.0333548448979855
10 0.0326308608055115
10 0.0293120518326759
10 0.0346559248864651
10 0.0344118997454643
10 0.0287452433258295
10 0.029734319075942
10 0.0320691652595997
10 0.035712044686079
10 0.0353247746825218
10 0.0317143984138966
10 0.0279940143227577
10 0.0289728082716465
10 0.0280836038291454
10 0.0274505820125341
10 0.032873272895813
10 0.0408668555319309
10 0.0284587554633617
10 0.0285261981189251
10 0.0347415171563625
10 0.0350285246968269
10 0.0282685402780771
10 0.0468988120555878
10 0.0283050537109375
10 0.0313708260655403
10 0.0344045981764793
10 0.050992090255022
10 0.0277141928672791
10 0.0305184964090586
};
\addplot [black]
table {%
10.75 0.900792926549911
11.25 0.900792926549911
11.25 0.92107692360878
10.75 0.92107692360878
10.75 0.900792926549911
};
\addplot [black]
table {%
11 0.900792926549911
11 0.870430588722229
};
\addplot [black]
table {%
11 0.92107692360878
11 0.950761079788208
};
\addplot [black]
table {%
10.875 0.870430588722229
11.125 0.870430588722229
};
\addplot [black]
table {%
10.875 0.950761079788208
11.125 0.950761079788208
};
\addplot [black, mark=*, mark size=0.25, mark options={solid,fill opacity=0}, only marks]
table {%
11 0.852525949478149
11 0.860657334327698
11 0.844575166702271
11 0.864339053630829
11 0.852579653263092
11 0.866922676563263
11 0.858353495597839
11 0.863872289657593
11 0.84400337934494
11 0.865719616413116
11 0.852221131324768
11 0.865303874015808
11 0.861790180206299
11 0.953101634979248
11 0.95465761423111
11 0.95437079668045
11 0.960362792015076
11 0.956890881061554
11 0.958955824375153
11 0.953275740146637
11 0.953281223773956
11 0.954690217971802
11 0.951795518398285
11 0.952345252037048
11 0.961099863052368
11 0.957894325256348
};
\addplot [black]
table {%
11.75 0.928844258189201
12.25 0.928844258189201
12.25 0.94757716357708
11.75 0.94757716357708
11.75 0.928844258189201
};
\addplot [black]
table {%
12 0.928844258189201
12 0.901132464408875
};
\addplot [black]
table {%
12 0.94757716357708
12 0.974228620529175
};
\addplot [black]
table {%
11.875 0.901132464408875
12.125 0.901132464408875
};
\addplot [black]
table {%
11.875 0.974228620529175
12.125 0.974228620529175
};
\addplot [black, mark=*, mark size=0.25, mark options={solid,fill opacity=0}, only marks]
table {%
12 0.890202522277832
12 0.887700915336609
12 0.89787095785141
12 0.890501856803894
12 0.899456679821014
12 0.894287288188934
12 0.899384677410126
12 0.891561985015869
12 0.874987125396729
12 0.894370138645172
12 0.866753756999969
12 0.878331363201141
12 0.900242149829865
12 0.899059653282166
12 0.891283929347992
12 0.975847780704498
12 0.977485060691833
12 0.975702881813049
12 0.985600233078003
12 0.999121963977814
12 0.986512780189514
12 0.977356255054474
12 0.986002624034882
12 0.985437870025635
12 0.992173612117767
12 0.982257127761841
12 0.978373825550079
12 0.992715179920197
12 0.980066061019897
12 0.983523726463318
12 0.980064153671265
12 0.976636469364166
12 0.98313170671463
12 0.97843861579895
12 1.00041460990906
};
\addplot [black]
table {%
12.75 0.00234020879724994
13.25 0.00234020879724994
13.25 0.00317230663495138
12.75 0.00317230663495138
12.75 0.00234020879724994
};
\addplot [black]
table {%
13 0.00234020879724994
13 0.00135593581944704
};
\addplot [black]
table {%
13 0.00317230663495138
13 0.00440270313993096
};
\addplot [black]
table {%
12.875 0.00135593581944704
13.125 0.00135593581944704
};
\addplot [black]
table {%
12.875 0.00440270313993096
13.125 0.00440270313993096
};
\addplot [black, mark=*, mark size=0.25, mark options={solid,fill opacity=0}, only marks]
table {%
13 0.00467237876728177
13 0.00484682014212012
13 0.00635237758979201
13 0.00474139535799623
13 0.00481424108147621
13 0.0045265955850482
13 0.00478761037811637
13 0.0110207898542285
13 0.00490204012021422
13 0.00580711383372545
13 0.00889790803194046
13 0.00447114603593946
13 0.00521790608763695
13 0.00518365297466516
13 0.00581438420340419
13 0.00819866172969341
13 0.00478567229583859
13 0.00765135604888201
13 0.00487250415608287
13 0.00486068800091743
13 0.00462410459294915
13 0.00504250125959516
13 0.00555961253121495
13 0.00466576172038913
13 0.00600865762680769
13 0.00492116063833237
13 0.00447005033493042
13 0.00740780914202332
13 0.00625187205150723
13 0.00516764167696238
13 0.00492809992283583
13 0.00537480926141143
13 0.00715387472882867
13 0.00514295836910605
13 0.00450631091371179
13 0.0045925434678793
13 0.00576888862997293
13 0.00736838486045599
13 0.00465897237882018
13 0.0050110649317503
13 0.00480390153825283
13 0.00489183887839317
13 0.00479673687368631
13 0.00446323491632938
13 0.00571291567757726
13 0.00567421549931169
13 0.00523528968915343
13 0.00493220193311572
13 0.00443366775289178
13 0.00466495472937822
13 0.00607906235381961
13 0.00444058235734701
13 0.00864538922905922
13 0.00465884152799845
13 0.00526070455089211
13 0.00451893918216228
13 0.00978034175932407
13 0.00517240446060896
13 0.00461588567122817
13 0.00475338008254766
};
\addplot [black]
table {%
13.75 0.922692403197289
14.25 0.922692403197289
14.25 0.944850981235504
13.75 0.944850981235504
13.75 0.922692403197289
};
\addplot [black]
table {%
14 0.922692403197289
14 0.889706313610077
};
\addplot [black]
table {%
14 0.944850981235504
14 0.977836608886719
};
\addplot [black]
table {%
13.875 0.889706313610077
14.125 0.889706313610077
};
\addplot [black]
table {%
13.875 0.977836608886719
14.125 0.977836608886719
};
\addplot [black, mark=*, mark size=0.25, mark options={solid,fill opacity=0}, only marks]
table {%
14 0.888362646102905
14 0.848779439926147
14 0.87933075428009
14 0.885771751403809
14 0.877241313457489
14 0.888303816318512
14 0.871411919593811
14 0.867824196815491
14 0.887353122234344
14 0.879668414592743
14 0.864513099193573
14 0.886791050434113
14 0.982074856758118
14 0.98545503616333
14 0.985044062137604
14 0.980445086956024
14 0.979530572891235
14 0.981665015220642
14 0.978512465953827
14 0.98252147436142
};
\addplot [black]
table {%
14.75 0.974203556776047
15.25 0.974203556776047
15.25 0.989086002111435
14.75 0.989086002111435
14.75 0.974203556776047
};
\addplot [black]
table {%
15 0.974203556776047
15 0.952045917510986
};
\addplot [black]
table {%
15 0.989086002111435
15 1.01140666007996
};
\addplot [black]
table {%
14.875 0.952045917510986
15.125 0.952045917510986
};
\addplot [black]
table {%
14.875 1.01140666007996
15.125 1.01140666007996
};
\addplot [black, mark=*, mark size=0.25, mark options={solid,fill opacity=0}, only marks]
table {%
15 0.938622653484344
15 0.947632372379303
15 0.951185047626495
15 0.942680418491364
15 0.936563789844513
15 0.951681315898895
15 0.945980846881866
15 1.01417207717896
15 1.01718032360077
15 1.01231682300568
15 1.01198959350586
15 1.01256966590881
15 1.01192104816437
15 1.01448547840118
15 1.06844854354858
15 1.01405274868011
15 1.01207506656647
15 1.01780235767365
};
\addplot [black]
table {%
15.75 0.00258085579844192
16.25 0.00258085579844192
16.25 0.00333802029490471
15.75 0.00333802029490471
15.75 0.00258085579844192
};
\addplot [black]
table {%
16 0.00258085579844192
16 0.00153939833398908
};
\addplot [black]
table {%
16 0.00333802029490471
16 0.0044667711481452
};
\addplot [black]
table {%
15.875 0.00153939833398908
16.125 0.00153939833398908
};
\addplot [black]
table {%
15.875 0.0044667711481452
16.125 0.0044667711481452
};
\addplot [black, mark=*, mark size=0.25, mark options={solid,fill opacity=0}, only marks]
table {%
16 0.00468183727934957
16 0.00531249959021807
16 0.0103612877428532
16 0.00709694065153599
16 0.00531616387888789
16 0.00575631484389305
16 0.00625986279919744
16 0.00711123226210475
16 0.0047905882820487
16 0.00607080105692148
16 0.005095140542835
16 0.00515897525474429
16 0.00596461910754442
16 0.0050191069021821
16 0.00490517821162939
16 0.00722440658137202
16 0.00455790106207132
16 0.00588459149003029
16 0.00468034204095602
16 0.00480867736041546
16 0.00556365679949522
16 0.00514072459191084
16 0.0060559599660337
16 0.00451266672462225
16 0.00469752447679639
16 0.0056651015765965
16 0.00658536283299327
16 0.00447787158191204
16 0.00453410018235445
16 0.00541446963325143
16 0.00484807277098298
16 0.00587406428530812
16 0.0045010126195848
16 0.00787585042417049
16 0.00450811162590981
16 0.00492102419957519
16 0.00907305721193552
16 0.00480131572112441
16 0.00450968323275447
};
\addplot [line width=1.5pt, red]
table {%
0.75 0.93349352478981
1.25 0.93349352478981
};
\addplot [line width=1.5pt, red]
table {%
1.75 0.926019728183746
2.25 0.926019728183746
};
\addplot [line width=1.5pt, red]
table {%
2.75 0.98372608423233
3.25 0.98372608423233
};
\addplot [line width=1.5pt, red]
table {%
3.75 0.926981151103973
4.25 0.926981151103973
};
\addplot [line width=1.5pt, red]
table {%
4.75 0.837072134017944
5.25 0.837072134017944
};
\addplot [line width=1.5pt, red]
table {%
5.75 0.0147622926160693
6.25 0.0147622926160693
};
\addplot [line width=1.5pt, red]
table {%
6.75 0.0294824754819274
7.25 0.0294824754819274
};
\addplot [line width=1.5pt, red]
table {%
7.75 0.0104186017997563
8.25 0.0104186017997563
};
\addplot [line width=1.5pt, red]
table {%
8.75 0.965815782546997
9.25 0.965815782546997
};
\addplot [line width=1.5pt, red]
table {%
9.75 0.0180582907050848
10.25 0.0180582907050848
};
\addplot [line width=1.5pt, red]
table {%
10.75 0.911589324474335
11.25 0.911589324474335
};
\addplot [line width=1.5pt, red]
table {%
11.75 0.938279539346695
12.25 0.938279539346695
};
\addplot [line width=1.5pt, red]
table {%
12.75 0.00269072246737778
13.25 0.00269072246737778
};
\addplot [line width=1.5pt, red]
table {%
13.75 0.934002488851547
14.25 0.934002488851547
};
\addplot [line width=1.5pt, red]
table {%
14.75 0.981973111629486
15.25 0.981973111629486
};
\addplot [line width=1.5pt, red]
table {%
15.75 0.00289832951966673
16.25 0.00289832951966673
};
\end{groupplot}

\end{tikzpicture}

%% file: main_icml.bbl
\begin{thebibliography}{45}
\providecommand{\natexlab}[1]{#1}
\providecommand{\url}[1]{\texttt{#1}}
\expandafter\ifx\csname urlstyle\endcsname\relax
  \providecommand{\doi}[1]{doi: #1}\else
  \providecommand{\doi}{doi: \begingroup \urlstyle{rm}\Url}\fi

\bibitem[Accardi(2001)]{accardi2001finetti}
Accardi, L.
\newblock De {F}inetti {T}heorem.
\newblock \emph{Hazewinkel, Michiel, Encyclopaedia of Mathematics, Kluwer
  Academic Publishers}, 2001.

\bibitem[Bauer \& Mnih(2019)Bauer and Mnih]{bauer2019resampled}
Bauer, M. and Mnih, A.
\newblock \capitalisewords{Resampled priors for variational autoencoders}.
\newblock In \emph{The 22nd International Conference on Artificial Intelligence
  and Statistics}, pp.\  66--75. PMLR, 2019.

\bibitem[Bengio et~al.(2013)Bengio, Courville, and
  Vincent]{bengio2013replearning}
Bengio, Y., Courville, A.~C., and Vincent, P.
\newblock \capitalisewords{Representation Learning: {A} Review and New
  Perspectives}.
\newblock \emph{{IEEE} Trans. Pattern Anal. Mach. Intell.}, 35\penalty0
  (8):\penalty0 1798--1828, 2013.
\newblock \doi{10.1109/TPAMI.2013.50}.
\newblock URL \url{https://doi.org/10.1109/TPAMI.2013.50}.

\bibitem[Bowman et~al.(2015)Bowman, Vilnis, Vinyals, Dai, Jozefowicz, and
  Bengio]{bowman2015generating}
Bowman, S.~R., Vilnis, L., Vinyals, O., Dai, A.~M., Jozefowicz, R., and Bengio,
  S.
\newblock \capitalisewords{Generating sentences from a continuous space}.
\newblock \emph{arXiv preprint arXiv:1511.06349}, 2015.

\bibitem[Burda et~al.(2016)Burda, Grosse, and Salakhutdinov]{burda2016iwae}
Burda, Y., Grosse, R.~B., and Salakhutdinov, R.
\newblock \capitalisewords{Importance Weighted Autoencoders}.
\newblock In Bengio, Y. and LeCun, Y. (eds.), \emph{4th International
  Conference on Learning Representations, {ICLR} 2016, San Juan, Puerto Rico,
  May 2-4, 2016, Conference Track Proceedings}, 2016.
\newblock URL \url{http://arxiv.org/abs/1509.00519}.

\bibitem[Chen et~al.(2017)Chen, Kingma, Salimans, Duan, Dhariwal, Schulman,
  Sutskever, and Abbeel]{chen2017postcollapse}
Chen, X., Kingma, D.~P., Salimans, T., Duan, Y., Dhariwal, P., Schulman, J.,
  Sutskever, I., and Abbeel, P.
\newblock \capitalisewords{Variational Lossy Autoencoder}.
\newblock In \emph{5th International Conference on Learning Representations,
  {ICLR} 2017, Toulon, France, April 24-26, 2017, Conference Track
  Proceedings}. OpenReview.net, 2017.
\newblock URL \url{https://openreview.net/forum?id=BysvGP5ee}.

\bibitem[Cremer et~al.(2017)Cremer, Morris, and
  Duvenaud]{cremer2017reinterpretingiwae}
Cremer, C., Morris, Q., and Duvenaud, D.
\newblock \capitalisewords{Reinterpreting Importance-Weighted Autoencoders}.
\newblock In \emph{5th International Conference on Learning Representations,
  {ICLR} 2017, Toulon, France, April 24-26, 2017, Workshop Track Proceedings}.
  OpenReview.net, 2017.
\newblock URL \url{https://openreview.net/forum?id=Syw2ZgrFx}.

\bibitem[Dai \& Wipf(2019)Dai and Wipf]{dai2018diagnosing}
Dai, B. and Wipf, D.
\newblock \capitalisewords{Diagnosing and Enhancing {VAE} Models}.
\newblock In \emph{International Conference on Learning Representations}, 2019.
\newblock URL \url{https://openreview.net/forum?id=B1e0X3C9tQ}.

\bibitem[Dai et~al.(2018)Dai, Wang, Aston, Hua, and Wipf]{dai2018connections}
Dai, B., Wang, Y., Aston, J., Hua, G., and Wipf, D.
\newblock \capitalisewords{Connections with robust PCA and the role of emergent
  sparsity in variational autoencoder models}.
\newblock \emph{The Journal of Machine Learning Research}, 19\penalty0
  (1):\penalty0 1573--1614, 2018.

\bibitem[Dai et~al.(2020)Dai, Wang, and Wipf]{dai2020usual}
Dai, B., Wang, Z., and Wipf, D.
\newblock \capitalisewords{The usual suspects? Reassessing blame for VAE
  posterior collapse}.
\newblock In \emph{International Conference on Machine Learning}, pp.\
  2313--2322. PMLR, 2020.

\bibitem[Doucet \& Johansen(2011)Doucet and Johansen]{Doucet2008ATO}
Doucet, A. and Johansen, A.~M.
\newblock \capitalisewords{A Tutorial on Particle Filtering and Smoothing:
  Fifteen years later}.
\newblock \emph{Oxford Handbook of Nonlinear Filtering}, 2011.

\bibitem[Ghosh et~al.(2019)Ghosh, Sajjadi, Vergari, Black, and
  Sch{\"o}lkopf]{ghosh2019variational}
Ghosh, P., Sajjadi, M.~S., Vergari, A., Black, M., and Sch{\"o}lkopf, B.
\newblock \capitalisewords{From variational to deterministic autoencoders}.
\newblock \emph{arXiv preprint arXiv:1903.12436}, 2019.

\bibitem[Goodfellow et~al.(2014)Goodfellow, Pouget{-}Abadie, Mirza, Xu,
  Warde{-}Farley, Ozair, Courville, and Bengio]{goodfellow2014gan}
Goodfellow, I.~J., Pouget{-}Abadie, J., Mirza, M., Xu, B., Warde{-}Farley, D.,
  Ozair, S., Courville, A.~C., and Bengio, Y.
\newblock \capitalisewords{Generative Adversarial Networks}.
\newblock \emph{CoRR}, abs/1406.2661, 2014.
\newblock URL \url{http://arxiv.org/abs/1406.2661}.

\bibitem[Heusel et~al.(2017)Heusel, Ramsauer, Unterthiner, Nessler, Klambauer,
  and Hochreiter]{heusel2017gans}
Heusel, M., Ramsauer, H., Unterthiner, T., Nessler, B., Klambauer, G., and
  Hochreiter, S.
\newblock \capitalisewords{GANs trained by a two time-scale update rule
  converge to a nash equilibrium}.
\newblock \emph{arXiv preprint arXiv:1706.08500}, 12\penalty0 (1), 2017.

\bibitem[Higgins et~al.(2017)Higgins, Matthey, Pal, Burgess, Glorot, Botvinick,
  Mohamed, and Lerchner]{higgins2017betavae}
Higgins, I., Matthey, L., Pal, A., Burgess, C.~P., Glorot, X., Botvinick,
  M.~M., Mohamed, S., and Lerchner, A.
\newblock \capitalisewords{beta-VAE: Learning Basic Visual Concepts with a
  Constrained Variational Framework}.
\newblock In \emph{5th International Conference on Learning Representations,
  {ICLR} 2017, Toulon, France, April 24-26, 2017, Conference Track
  Proceedings}. OpenReview.net, 2017.
\newblock URL \url{https://openreview.net/forum?id=Sy2fzU9gl}.

\bibitem[Hoffman \& Johnson(2016)Hoffman and Johnson]{hoffman2016elbo}
Hoffman, M.~D. and Johnson, M.~J.
\newblock \capitalisewords{ELBO surgery: Yet another way to carve up the
  evidence lower bound}.
\newblock In \emph{Proc. Workshop Adv. Approx. Bayesian Inference}, pp.\ ~2,
  2016.

\bibitem[Julier et~al.(2000)Julier, Uhlmann, and Durrant-Whyte]{julier2000new}
Julier, S., Uhlmann, J., and Durrant-Whyte, H.~F.
\newblock \capitalisewords{A new method for the nonlinear transformation of
  means and covariances in filters and estimators}.
\newblock \emph{IEEE Transactions on automatic control}, 45\penalty0
  (3):\penalty0 477--482, 2000.

\bibitem[Karl et~al.(2016)Karl, Soelch, Bayer, and Van~der Smagt]{karl2016deep}
Karl, M., Soelch, M., Bayer, J., and Van~der Smagt, P.
\newblock \capitalisewords{Deep variational bayes filters: Unsupervised
  learning of state space models from raw data}.
\newblock \emph{arXiv preprint arXiv:1605.06432}, 2016.

\bibitem[Khromov \& Singh(2023)Khromov and Singh]{khromov2023fundamental}
Khromov, G. and Singh, S.~P.
\newblock Some fundamental aspects about lipschitz continuity of neural network
  functions, 2023.

\bibitem[Kingma \& Welling(2013)Kingma and Welling]{kingma2013auto}
Kingma, D.~P. and Welling, M.
\newblock \capitalisewords{Auto-encoding variational bayes}.
\newblock \emph{arXiv preprint arXiv:1312.6114}, 2013.

\bibitem[Kingma et~al.(2015)Kingma, Salimans, and
  Welling]{kingma2015variational}
Kingma, D.~P., Salimans, T., and Welling, M.
\newblock \capitalisewords{Variational dropout and the local reparameterization
  trick}.
\newblock \emph{Advances in neural information processing systems}, 28, 2015.

\bibitem[Krizhevsky et~al.(2009)Krizhevsky, Hinton,
  et~al.]{krizhevsky2009learning}
Krizhevsky, A., Hinton, G., et~al.
\newblock \capitalisewords{Learning multiple layers of features from tiny
  images}.
\newblock 2009.

\bibitem[Kusner et~al.(2017)Kusner, Paige, and
  Hern{\'a}ndez-Lobato]{kusner2017grammar}
Kusner, M.~J., Paige, B., and Hern{\'a}ndez-Lobato, J.~M.
\newblock \capitalisewords{Grammar Variational Autoencoder}.
\newblock In Precup, D. and Teh, Y.~W. (eds.), \emph{Proceedings of the 34th
  International Conference on Machine Learning}, volume~70 of \emph{Proceedings
  of Machine Learning Research}, pp.\  1945--1954. PMLR, 06--11 Aug 2017.
\newblock URL \url{https://proceedings.mlr.press/v70/kusner17a.html}.

\bibitem[Liu et~al.(2015)Liu, Luo, Wang, and Tang]{liu2015faceattributes}
Liu, Z., Luo, P., Wang, X., and Tang, X.
\newblock \capitalisewords{Deep Learning Face Attributes in the Wild}.
\newblock In \emph{Proceedings of International Conference on Computer Vision
  (ICCV)}, December 2015.

\bibitem[Lucas et~al.(2019)Lucas, Tucker, Grosse, and
  Norouzi]{lucas2019postcollapse}
Lucas, J., Tucker, G., Grosse, R.~B., and Norouzi, M.
\newblock \capitalisewords{Understanding Posterior Collapse in Generative
  Latent Variable Models}.
\newblock In \emph{Deep Generative Models for Highly Structured Data, {ICLR}
  2019 Workshop, New Orleans, Louisiana, United States, May 6, 2019}.
  OpenReview.net, 2019.
\newblock URL \url{https://openreview.net/forum?id=r1xaVLUYuE}.

\bibitem[Mathieu et~al.(2019)Mathieu, Rainforth, Siddharth, and
  Teh]{mathieu2019disentangling}
Mathieu, E., Rainforth, T., Siddharth, N., and Teh, Y.~W.
\newblock \capitalisewords{Disentangling Disentanglement in Variational
  Autoencoders}.
\newblock In Chaudhuri, K. and Salakhutdinov, R. (eds.), \emph{Proceedings of
  the 36th International Conference on Machine Learning}, volume~97 of
  \emph{Proceedings of Machine Learning Research}, pp.\  4402--4412. PMLR,
  09--15 Jun 2019.
\newblock URL \url{https://proceedings.mlr.press/v97/mathieu19a.html}.

\bibitem[Menegaz et~al.(2015)Menegaz, Ishihara, Borges, and
  Vargas]{menegaz2015systematization}
Menegaz, H.~M., Ishihara, J.~Y., Borges, G.~A., and Vargas, A.~N.
\newblock \capitalisewords{A systematization of the unscented Kalman filter
  theory}.
\newblock \emph{IEEE Transactions on automatic control}, 60\penalty0
  (10):\penalty0 2583--2598, 2015.

\bibitem[Paszke et~al.(2019)Paszke, Gross, Massa, Lerer, Bradbury, Chanan,
  Killeen, Lin, Gimelshein, Antiga, et~al.]{paszke2019pytorch}
Paszke, A., Gross, S., Massa, F., Lerer, A., Bradbury, J., Chanan, G., Killeen,
  T., Lin, Z., Gimelshein, N., Antiga, L., et~al.
\newblock \capitalisewords{Pytorch: An imperative style, high-performance deep
  learning library}.
\newblock \emph{Advances in neural information processing systems}, 32, 2019.

\bibitem[Patrini et~al.(2020)Patrini, van~den Berg, Forre, Carioni, Bhargav,
  Welling, Genewein, and Nielsen]{patrini2020sinkhorn}
Patrini, G., van~den Berg, R., Forre, P., Carioni, M., Bhargav, S., Welling,
  M., Genewein, T., and Nielsen, F.
\newblock \capitalisewords{Sinkhorn autoencoders}.
\newblock In \emph{Uncertainty in Artificial Intelligence}, pp.\  733--743.
  PMLR, 2020.

\bibitem[Rainforth et~al.(2018)Rainforth, Kosiorek, Le, Maddison, Igl, Wood,
  and Teh]{rainforth2018tighter}
Rainforth, T., Kosiorek, A., Le, T.~A., Maddison, C., Igl, M., Wood, F., and
  Teh, Y.~W.
\newblock \capitalisewords{Tighter variational bounds are not necessarily
  better}.
\newblock In \emph{International Conference on Machine Learning}, pp.\
  4277--4285. PMLR, 2018.

\bibitem[Rezende et~al.(2014)Rezende, Mohamed, and
  Wierstra]{rezende2014variational}
Rezende, D.~J., Mohamed, S., and Wierstra, D.
\newblock \capitalisewords{Stochastic Backpropagation and Approximate Inference
  in Deep Generative Models}.
\newblock In \emph{Proceedings of the 31st International Conference on
  International Conference on Machine Learning - Volume 32}, ICML'14, pp.\
  II–1278–II–1286. JMLR.org, 2014.

\bibitem[Rolinek et~al.(2019)Rolinek, Zietlow, and
  Martius]{rolinek2019variational}
Rolinek, M., Zietlow, D., and Martius, G.
\newblock \capitalisewords{Variational autoencoders pursue PCA directions (by
  accident)}.
\newblock In \emph{Proceedings of the IEEE/CVF Conference on Computer Vision
  and Pattern Recognition}, pp.\  12406--12415, 2019.

\bibitem[Rombach et~al.(2022)Rombach, Blattmann, Lorenz, Esser, and
  Ommer]{rombach2022high}
Rombach, R., Blattmann, A., Lorenz, D., Esser, P., and Ommer, B.
\newblock \capitalisewords{High-resolution image synthesis with latent
  diffusion models}.
\newblock In \emph{Proceedings of the IEEE/CVF Conference on Computer Vision
  and Pattern Recognition}, pp.\  10684--10695, 2022.

\bibitem[Seitzer(2020)]{Seitzer2020FID}
Seitzer, M.
\newblock \capitalisewords{{pytorch-fid: FID Score for PyTorch}}.
\newblock \url{https://github.com/mseitzer/pytorch-fid}, August 2020.
\newblock Version 0.2.1.

\bibitem[Tolstikhin et~al.(2018)Tolstikhin, Bousquet, Gelly, and
  Sch{\"{o}}lkopf]{tolstikhin2018wae}
Tolstikhin, I.~O., Bousquet, O., Gelly, S., and Sch{\"{o}}lkopf, B.
\newblock \capitalisewords{Wasserstein Auto-Encoders}.
\newblock In \emph{6th International Conference on Learning Representations,
  {ICLR} 2018, Vancouver, BC, Canada, April 30 - May 3, 2018, Conference Track
  Proceedings}. OpenReview.net, 2018.
\newblock URL \url{https://openreview.net/forum?id=HkL7n1-0b}.

\bibitem[Townsend et~al.(2019)Townsend, Bird, and
  Barber]{townsend2019compression}
Townsend, J., Bird, T., and Barber, D.
\newblock \capitalisewords{Practical lossless compression with latent variables
  using bits back coding}.
\newblock In \emph{7th International Conference on Learning Representations,
  {ICLR} 2019, New Orleans, LA, USA, May 6-9, 2019}. OpenReview.net, 2019.
\newblock URL \url{https://openreview.net/forum?id=ryE98iR5tm}.

\bibitem[Tripp et~al.(2020)Tripp, Daxberger, and
  Hern{\'{a}}ndez{-}Lobato]{tripp2020lso}
Tripp, A., Daxberger, E.~A., and Hern{\'{a}}ndez{-}Lobato, J.~M.
\newblock \capitalisewords{Sample-Efficient Optimization in the Latent Space of
  Deep Generative Models via Weighted Retraining}.
\newblock In Larochelle, H., Ranzato, M., Hadsell, R., Balcan, M., and Lin, H.
  (eds.), \emph{Advances in Neural Information Processing Systems 33: Annual
  Conference on Neural Information Processing Systems 2020, NeurIPS 2020,
  December 6-12, 2020, virtual}, 2020.

\bibitem[Tucker et~al.(2018)Tucker, Lawson, Gu, and Maddison]{tucker2018doubly}
Tucker, G., Lawson, D., Gu, S., and Maddison, C.~J.
\newblock \capitalisewords{Doubly reparameterized gradient estimators for monte
  carlo objectives}.
\newblock \emph{arXiv preprint arXiv:1810.04152}, 2018.

\bibitem[Uhlmann(1995)]{uhlmann1995dynamic}
Uhlmann, J.
\newblock \emph{\capitalisewords{Dynamic map building and localization: new
  theoretical foundations.}}
\newblock PhD thesis, University of Oxford, 1995.

\bibitem[Vahdat \& Kautz(2020)Vahdat and Kautz]{vahdat2020NVAE}
Vahdat, A. and Kautz, J.
\newblock \capitalisewords{{NVAE}: A Deep Hierarchical Variational
  Autoencoder}.
\newblock In \emph{Neural Information Processing Systems (NeurIPS)}, 2020.

\bibitem[Xiao et~al.(2017)Xiao, Rasul, and Vollgraf]{xiao2017fashion}
Xiao, H., Rasul, K., and Vollgraf, R.
\newblock \capitalisewords{Fashion-MNIST: a novel image dataset for
  benchmarking machine learning algorithms}.
\newblock \emph{arXiv preprint arXiv:1708.07747}, 2017.

\bibitem[Xu et~al.(2015)Xu, Wang, Chen, and Li]{xu2015empirical}
Xu, B., Wang, N., Chen, T., and Li, M.
\newblock \capitalisewords{Empirical evaluation of rectified activations in
  convolutional network}.
\newblock \emph{arXiv preprint arXiv:1505.00853}, 2015.

\bibitem[Zhang et~al.(2018)Zhang, B{\"u}tepage, Kjellstr{\"o}m, and
  Mandt]{zhang2018advances}
Zhang, C., B{\"u}tepage, J., Kjellstr{\"o}m, H., and Mandt, S.
\newblock \capitalisewords{Advances in variational inference}.
\newblock \emph{IEEE transactions on pattern analysis and machine
  intelligence}, 41\penalty0 (8):\penalty0 2008--2026, 2018.

\bibitem[Zhang et~al.(2009)Zhang, Liu, and Zhao]{zhang2009accuracy}
Zhang, W., Liu, M., and Zhao, Z.-g.
\newblock \capitalisewords{Accuracy analysis of unscented transformation of
  several sampling strategies}.
\newblock In \emph{2009 10th ACIS International Conference on Software
  Engineering, Artificial Intelligences, Networking and Parallel/Distributed
  Computing}, pp.\  377--380. IEEE, 2009.

\bibitem[Zietlow et~al.(2021)Zietlow, Rolinek, and
  Martius]{zietlow2021demystifying}
Zietlow, D., Rolinek, M., and Martius, G.
\newblock \capitalisewords{Demystifying Inductive Biases for (Beta-) VAE Based
  Architectures}.
\newblock In \emph{International Conference on Machine Learning}, pp.\
  12945--12954. PMLR, 2021.

\end{thebibliography}
